\documentclass[10pt, journal]{IEEEtran}
\IEEEoverridecommandlockouts 

\usepackage{amsmath,amssymb,amsfonts}
\usepackage{graphicx, stfloats}
\usepackage{pifont}

\usepackage{booktabs}
\usepackage{amsthm}
\usepackage{xcolor}
\usepackage[table]{xcolor}
\usepackage[figuresright]{rotating}

\graphicspath{{/}{fig/}}

\usepackage{array}
\usepackage{textcomp}
\usepackage{multirow}

\usepackage{mathtools}
\usepackage{breqn}
\usepackage{float}

\usepackage{tikz}
\usetikzlibrary{arrows.meta,positioning,fit,backgrounds,calc}
\usepackage{pgfplots}
\pgfplotsset{compat=1.7}
\usepgfplotslibrary{groupplots}

\usepackage{caption}
\usepackage{subcaption}

\usepackage{tabularx}
\usepackage{hyperref}
\usepackage{flushend}
\usepackage{algorithmic}
\usepackage[vlined, ruled, shortend]{algorithm2e}

\usepackage{multicol}
\usepackage{times}
\usepackage{cleveref}
\usepackage{soul}

\usepackage{cuted}
\usepackage{etoolbox}
\usepackage{makecell}
\usepackage{enumitem}

\definecolor{jColor}{HTML}{00A676}

\newlength\figureheight
\newlength\figurewidth
\SetAlCapNameFnt{\footnotesize}
\SetAlCapFnt{\footnotesize}

\newcommand{\ie}{\textit{i.e.,}}%
\newcommand{\eg}{\textit{e.g.,}}%
\newcommand{\etal}{\textit{et al.}}%

\crefname{figure}{Fig.}{Figs.}
\Crefname{figure}{Fig.}{Figs.}
\Crefname{section}{Section}{Section}
\crefname{section}{Section}{Section}
\crefname{equation}{Eq.}{eq.}
\Crefname{equation}{Eq.}{eq.}

\title{
    Kairos: Grounded Forecasting of Presence and Directional Flow \\ in 4D Scene Graphs
}

\author{
    Iacopo Catalano, Julio A. Placed, Javier Civera, Jorge Peña-Queralta
    \thanks{This work was partially supported by the Finnish Cultural Foundation, and by DGA\_FSE T73\_23R.
    I. Catalano is with the University of Turku, Finland.
    J. Pe\~na-Queralta is with the Centre for Artificial Intelligence, Zürich University of Applied Sciences, Winterthur, Switzerland.
    J. A.~Placed is with the Instituto Tecnol\'ogico de Arag\'on (ITA) and the University of Zaragoza, Spain.
    J. Civera is with the University of Zaragoza, Spain.
    Corresponding: \texttt{\small imcata@utu.fi}
    }
}

\begin{document}

\maketitle

\begin{abstract}%
    \label{sec:abstract}
    Long-term autonomy in human-populated environments requires anticipating whether and how people will move at times a robot has not yet observed. Existing representations of pedestrian motion face a tradeoff: they either forecast future activity, reducing each location to a scalar rate, or model the full directional distribution, holding it fixed in time. We present Kairos, a predictive directional-flow memory that extends a hierarchical 3D scene graph (3DSG) to a 4D scene graph (4DSG). Every observed voxel of the reconstructed geometry stores a directional mixture and a presence rate, and spectral predictors forecast, for any future query time, both the probability that people are present and the full directional distribution of their motion. Pairwise flow dependence between adjacent voxels supports conditional queries, and per-voxel predictive variances yield calibrated credible intervals that tighten as observations accumulate. We evaluate Kairos on three real pedestrian environments: a robot-collected campus dataset, a shopping mall, and a station concourse recorded continuously for eleven months. Its learned state remains consistent under loop-closure corrections, and its forecasts are competitive with dedicated occupancy and flow models trained on the full detection stream, although Kairos learns from only the small fraction available to a patrolling robot. Finally, we validate the representation on a downstream encounter-probability planning task, where plans computed over the Kairos forecasts encounter more people than plans computed over any time-invariant map at an equal success rate. We provide the code at \footnotesize{\url{https://github.com/IacopomC/kairos}}
\end{abstract}

\IEEEpeerreviewmaketitle

\section{Introduction}\label{sec:introduction}

\begin{figure}[t]
    \centering
    \includegraphics[width=\linewidth]{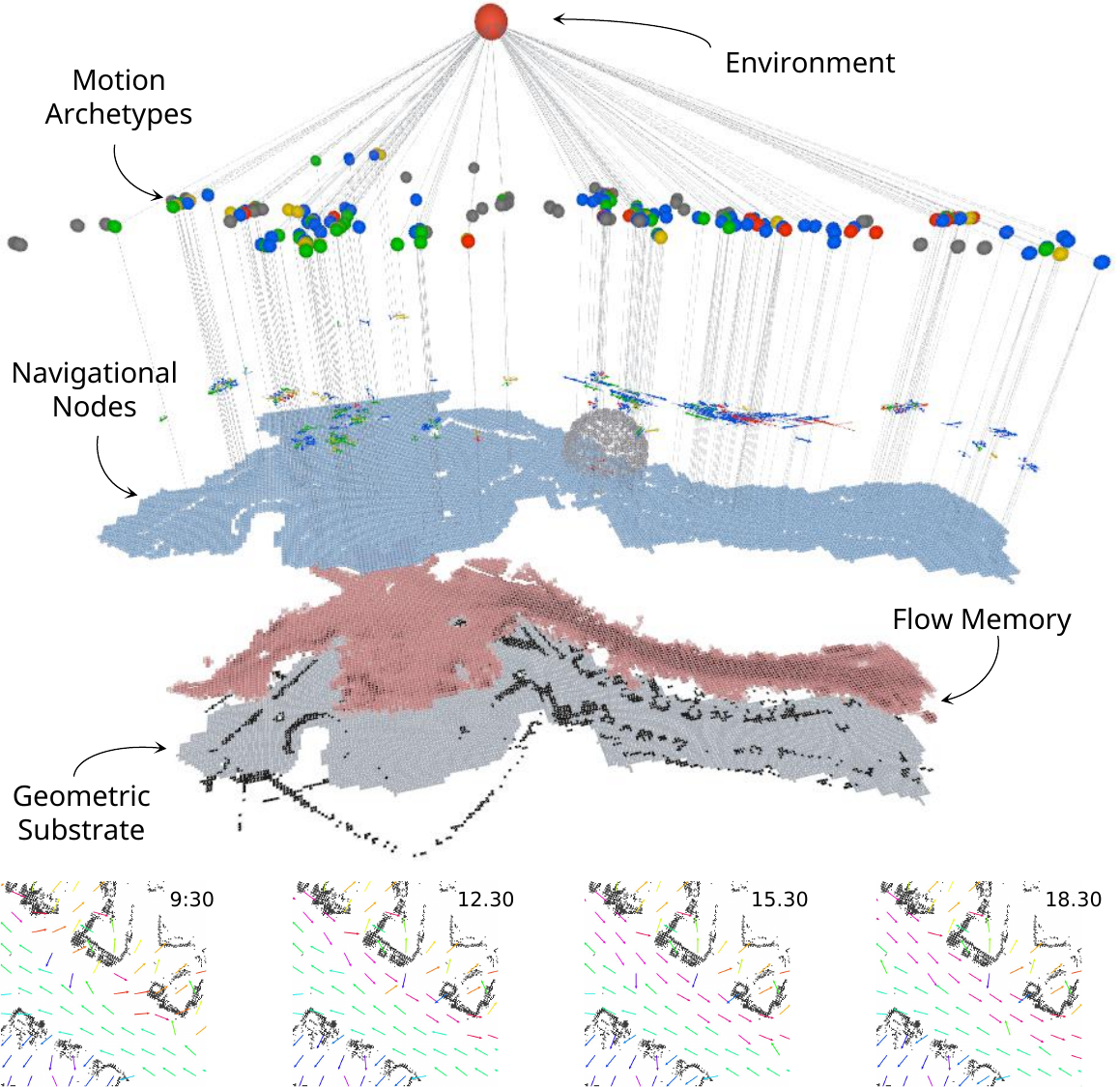}
    \caption{\textbf{Top}: 4D scene graph after a one-day exploration in ATC~\cite{brvsvcic2013person}. The geometric gray substrate holds the occupancy map, with occupied cells in black, and flow voxels anchored to it in red, shaded by accumulated detections. The navigational layer on top of them carries the lifted flow: each arrow is a mixture component of the predicted heading distribution at that place, colored by its motion archetype (gray static, green unimodal, blue bimodal, red multimodal, yellow diffuse). Archetype nodes group adjacent navigational nodes of one archetype under the environment node. The translucent sphere marks the robot's active sensing window. \textbf{Bottom}: Predicted flow over a portion of the east corridor at four times of the same day. Colors encode heading on a circular scale.
    }
    \label{fig:4dsg}
\end{figure}

Environments shared with people are shaped by human motion. Long-term autonomy in such environments requires anticipating how people move through a space at a given time, a structure that cannot be inferred from observations of visible agents alone. Pedestrian flow is neither static nor random. For example, a corridor may carry predominantly one-way traffic during the morning commute and reverse later in the day, while a station platform may empty outward after a train arrives and fill again inward before the next departure. Such patterns are structured by daily and weekly periodicities that can be recovered from the sparse, irregular observations a robot gathers over long-term deployment~\cite{krajnik2017fremen, jovan2016poisson}. To support predictive planning, a flow model must be \emph{directional} (a multi-modal distribution over heading, not just occupancy), \emph{time-conditioned} (a distribution that shifts with the hour), and \emph{grounded} (a distribution tied to the geometry the robot reconstructs and plans on).

\emph{Maps of Dynamics} (MoDs) are models of motion learned as a persistent property of the environment~\cite{kucner2023survey}. Spectral occupancy models~\cite{krajnik2017fremen, krajnik2014froctomap} forecast in time but predict only a scalar per cell, indicating whether a location will be busy but without describing how motion through it is organized. Directional models represent the full motion distribution but only as a long-run average, fixed once estimated and blind to how it shifts over the day~\cite{kucner2013conditional, kucner2017enabling, senanayake2018directional}.

\emph{Hierarchical 3D Scene Graphs} (3DSG) provide a semantic structure of the environment grounded in geometry~\cite{catalano20253d} but their temporal dimension tracks individual agents~\cite{rosinol2021kimera, greve2023curb}, state changes~\cite{looper20233d}, or the scene's history for retrospective query~\cite{Schmid-RSS24-Khronos, gorlo2025describe}, capturing what happened rather than the recurring flow of a location. Recent work instead embeds a directional motion model as a layer of the hierarchy, fitting a per-node semi-wrapped Gaussian mixture~\cite{catalano2025rheos}, or forecasting directional flow over discrete orientation bins~\cite{catalano2025aion}. However, in both cases the estimate is bound to navigational nodes by proximity rather than to the reconstructed geometry.

Neither representation satisfies all three properties at once: a scalar state, a fixed average, a node-bound estimate, or per-agent tracks cannot, by construction, forecast how a location will flow at a future time.
We present \textbf{Kairos}, a \emph{predictive directional-flow memory} that extends a 3DSG to a 4D scene graph (4DSG).
A spectral predictor on each mixing weight of a per-voxel directional mixture forecasts the entire distribution over time, anchored to the observed voxels of the robot's TSDF reconstruction and kept consistent under map re-optimization.

\begin{table*}[t]
    \centering
    \caption{\textbf{Comparison of directional and spectral flow-dynamics methods}. The first five columns refer to properties that flow forecasting must satisfy. The last three are additional capabilities beyond the required set. --- stands for \emph{not provided}.}
    \label{tab:comparison}
    \scriptsize
    \setlength{\tabcolsep}{4pt}
    \begin{tabular*}{\textwidth}{@{\extracolsep{\fill}}ll ccccc ccc@{}}
        \toprule
        & & \multicolumn{5}{c}{\emph{Required properties}} & \multicolumn{3}{c}{\emph{Additional capabilities}} \\
        \cmidrule(lr){3-7}\cmidrule(lr){8-10}
        Family & Method & \makecell[c]{Flow\\Representation} & Forecast & Index & Warp & \makecell[c]{Online\\Fit} & \makecell[c]{Flow\\Coupling} & Presence & \makecell[c]{Forecast\\Uncertainty} \\
        \midrule
        \multirow{6}{*}{\makecell[c]{\emph{Discrete MoD}}}
        & Cond. Transition Maps~\cite{kucner2013conditional} & transitions & --- & grid & --- & --- & implicit & --- & ---\\
        & CLiFF-map~\cite{kucner2017enabling} & SW-GMM & --- & grid & --- & --- & kernel imp. & --- & ---\\
        & Dir. Grid Maps~\cite{senanayake2018directional} & von Mises mixture & --- & grid & --- & --- & --- & --- & ---\\
        & FreMEn~\cite{krajnik2017fremen}, FrOctoMap~\cite{krajnik2014froctomap} & --- & occupancy & grid & --- & \checkmark & --- & binary & ---\\
        & STeF-Map~\cite{molina2021robotic} & 8-bin histogram & bin value & grid & --- & \checkmark & --- & --- & ---\\
        & Online CLiFF-map~\cite{zhu2025cliffonline} & SW-GMM & --- & grid & --- & \checkmark & --- & --- & ---\\
        \cmidrule(l{0pt}r{0pt}){1-10}
        \multirow{2}{*}{\makecell[c]{\emph{Continuous MoD}}}
        & NeMo-map~\cite{zhu2025neural} & SW-GMM & mixture params. & coordinates & --- & --- & --- & --- & ---\\
        & Variational GP activity~\cite{stuede2022non} & --- & activity rate & coordinates & --- & --- & --- & GP rate & --- \\
        \cmidrule(l{0pt}r{0pt}){1-10}
        \multirow{3}{*}{\makecell[c]{\emph{Scene-graph Dynamics}}}
        & Aion~\cite{catalano2025aion} & 8-bin histogram & bin value & nodes & transfer & \checkmark & --- & --- & ---\\
        & Rheos~\cite{catalano2025rheos} & SW-GMM & --- & nodes & transfer & \checkmark & --- & --- & ---\\
        & {Kairos} \emph{(ours)} & SW-GMM, $K=8$ & mixing weights & voxels & re-key & \checkmark & pairwise & Gamma--Poisson & posterior\\
        \bottomrule
    \end{tabular*}
\end{table*}

At any traversable location and any future time, Kairos answers two questions a static map cannot: whether agents are likely to be present there, and if so, how they are likely to be moving, as a full directional distribution. A presence channel kept separate from flow distinguishes \emph{no one passes here} from \emph{no one has yet been observed}. Because the forecast covers the whole reconstructed environment and is exposed on the navigational graph, a robot can weigh entire routes before committing to one, comparing how each will flow at the time it would traverse it rather than planning against a snapshot.

We make the following contributions:
\begin{itemize}[leftmargin=*, nosep]
    \item \textbf{Full-distribution flow forecasting on the map}: 
    Kairos is the first method to forecast the complete directional flow distribution, where prior flow layers predict a scalar summary or a static average, and continuous full-distribution forecasts are learned offline over abstract coordinates.
    Each voxel is updated online from a fixed-size summary, at a cost independent of how long the robot has been observing.
    \item \textbf{Grounded under map correction}: the forecast is  \emph{anchored to the geometry the robot reconstructs online}, coindexed with the scene graph's own cells and equivariant under the corrections its backend applies, two properties that grid-based and proximity-attached flow maps do not satisfy.
    \item \textbf{Decoupled presence and flow}: a per-voxel presence channel distinguishes \emph{no agents present} from \emph{no flow}, so planners can reason independently about occupancy and motion.
    \item \textbf{Planner-facing integration}: per-voxel forecasts are aggregated up the hierarchy and annotated on the navigational graph, exposing time-conditioned flow through the standard scene-graph interface; we demonstrate it on a downstream encounter-probability planning task.
\end{itemize}

We further show that pairwise flow dependence over adjacent voxels supports conditional queries and joint draws coherent with continuous motion, and that combining the posterior variance of the spectral coefficients with the measured scatter of the observations returns every forecast as a calibrated credible interval.
\section{Related Work}\label{sec:related_work}

Kairos draws on two complementary lines of work: MoDs, which model motion as a persistent property of the environment~\cite{kucner2023survey}, and 3DSGs, which provide semantic and topological structure to environment representations~\cite{catalano20253d}. We review both against the three properties required for anticipatory planning: \emph{directionality}, \emph{time-conditioning}, and \emph{grounding}. See~\Cref{tab:comparison}.

\subsection{Maps of Dynamics}

\textbf{Directional models of flow.} Conditional transition maps~\cite{kucner2013conditional} 
model flow at each cell as a distribution over transitions to its neighbors, capturing dependencies between adjacent locations rather than estimating them independently. The resulting model is time-invariant.
CLiFF-map~\cite{kucner2017enabling} instead fits a per-cell Semi-Wrapped Gaussian Mixture Model (SW-GMM) over direction and speed, representing multimodal flow as a continuous density on the cylinder (the parametrization Kairos adopts).
Directional grid maps~\cite{senanayake2018directional} model angular uncertainty at each cell using a von Mises mixture, omitting speed. These approaches meet the directional requirement, but all estimate a long-run average at each cell, estimated once and fixed thereafter, with no temporal dependence.

\textbf{Spectral models of periodic state} represent the environment as a set of periodic processes recovered in the frequency domain from sparse, irregular observations~\cite{krajnik2014spectral}. FreMEn (Frequency Map Enhancement)~\cite{krajnik2017fremen} formalizes this for a binary state, retaining Fourier coefficients at candidate periods and evaluating them at any future time to obtain a probability. 
This representation is agnostic to what the state describes, which has indeed allowed its application to different domains.
FrOctoMap~\cite{krajnik2014froctomap} places FreMEn in a voxel grid for spatio-temporal occupancy; the Poisson-spectral model of~\cite{jovan2016poisson} replaces the binary state with a counting process to forecast activity rate; STeF-Map~\cite{molina2021robotic} applies FreMEn to per-cell orientation histograms.
Information-driven mapping refines such models online through an incremental per-cell spectral update~\cite{santos2015lifelong} (the form Kairos retains), and wrapped-time representations extend such spectral models to multi-day and non-sinusoidal periodicity~\cite{vintr2019time}.
These models meet the time-conditioned requirement, but each tracks a \emph{scalar} per cell: an occupancy state, a rate, or a single histogram bin. STeF-Map comes closest, yet a histogram is a quantized, per-bin scalar rather than a continuous directional distribution.
Kairos retains FreMEn spectral predictor and instantiates one per scalar channel of a distribution-valued state: on each mixing weight of a per-voxel SW-GMM, and on the occupancy rate.

\textbf{Learned models.} Recent work relaxes these limits along three axes. The first replaces the analytic spectral primitive with a learned temporal model. Neural sequence models predict how the motion field itself evolves~\cite{shi2025learning}; a hybrid model pairs a periodic spectral term with an LSTM residual over orientation histograms~\cite{yan2025stef}; and event-triggered formulations add adaptive cells and neural stochastic differential equations to track abrupt, non-stationary change~\cite{shi2025event}, which recurrence alone cannot represent.
The second relaxes the batch fit: online estimators update a per-cell mixture from streaming observations to follow drift without rebuilding~\cite{zhu2025cliffonline}. 
The third uses a continuous formulation, removing the grid altogether. NeMo-map~\cite{zhu2025neural} learns an implicit neural field from spatio-temporal coordinates to the parameters of an SW-GMM, so it outputs a directional distribution at any location and future time without discretizing either. However, it is fit offline in batch and its readout is a function of a coordinate rather than of an element of the built map.
A fourth direction of research focuses on estimating the field from scarce or absent local observations: a Gaussian process interpolates over sparse mobile-robot observations~\cite{stuede2022non} and outputs a variance with each estimate, while transfer approaches infer the field from a floor plan~\cite{verdoja2024bayesian} or regress it from egocentric views~\cite{catalano2026egomod}, providing a prior where it has not been initialized yet.

Individually the models above meet the directional and time-conditioned requirements, but none satisfies the grounding condition: each lives on an abstract grid or coordinate space, decoupled from the geometry a robot reconstructs and plans on, so the estimate cannot follow the map as it is corrected.
In contrast, Kairos keys each channel to the active voxels of the robot's reconstruction, and re-keys them after every correction the backend applies, so the forecast is defined on (and consistent with) the geometry the planner reads.

\subsection{Dynamics in 3D Scene Graphs}

3DSGs layer the world by increasing abstraction, grounding semantics and topology in the geometry the robot plans on, and thus satisfying by construction the one property the models above lack.
Most 3DSGs are quasi-static; where dynamics are modeled, they take three forms: interaction-driven state change, per-entity temporal dynamics, and aggregate flow.

\textbf{Interaction-driven state change.} Action-aware scene graphs annotate nodes with affordances but treat them as time-invariant~\cite{ravichandran2022hierarchical, agia2022taskography}. Looper \etal~\cite{looper20233d} attach a learned variability score that estimates how likely an object is to change, without specifying when. This family captures whether and how much a scene changes, but not when or how motion flows through it (\ie~a property of objects rather than the motion through a place). Therefore, neither the directional nor the time-conditioned requirements apply.

\textbf{Per-entity temporal dynamics.} Kimera-DSG~\cite{rosinol2021kimera} models moving agents as dynamic nodes and CURB-SG~\cite{greve2023curb} tracks vehicles for urban navigation. Khronos~\cite{Schmid-RSS24-Khronos} and DAAAM~\cite{gorlo2025describe} maintain geometrically grounded scene histories for retrospective queries, and~\cite{Schmid-RSS24-Khronos} in particular maintains a consistent representation under backend optimization. However, their temporal models are defined independently for each entity: they track the history of individual agents rather than estimating recurring patterns of motion. Consequently, they cannot represent the flow associated with a location, nor do they support queries about its expected state at future times.

\textbf{Aggregate flow as a graph layer.} Closest to Kairos, two systems attach a MoD to the navigational layer.
Aion~\cite{catalano2025aion} keeps a FreMEn model over each orientation histogram and forecasts three scalar quantities (magnitude, dominant direction, and entropy), discarding the distribution itself; Rheos~\cite{catalano2025rheos} keeps the full distribution as a continuous SW-GMM but fits it once, with no dependence on time.
Each therefore meets one of the two requirements the other misses.
Both associate observations to navigational nodes by proximity, rather than the reconstructed geometry. 
Kairos closes both gaps by forecasting the full directional distribution, and keying the estimate to the active voxels of the reconstruction rather than to node proximity, thus holding under the backend corrections.

\section{Problem Formulation}
\label{sec:problem}


We consider the long-term deployment of a robot in an environment populated with other mobile agents, \emph{e.g.}, humans. 
The robot observes a stream of detections $\mathcal{Z}^{M}_{\le T} = \{(t^{j}_k, \mathbf{x}^{j}_k, \mathbf{v}^{j}_k) : t^{j}_k \le T\}_{j=1}^{M}$  for $M$ agents up to time $T$, where, in the most general case, $\mathbf{x}^{j}_k \in \mathbb{R}^3_{}$ denotes the 3D position of an agent $j$ at time $t^{j}_k$, and $\mathbf{v}^{j}_k \in \mathbb{R}^3_{}$ its corresponding velocity. Assuming that agents move on the ground, we characterize the flow by the horizontal component of their velocity. Let $\Pi : \mathbb{R}^3_{} \to\mathbb{R}^2_{}$ denote the projection onto the horizontal plane, such that $\Pi\mathbf{v}^{j}_k = (v^{j}_{k,x}, v^{j}_{k,y})$. Each detection then defines a heading $\theta^{j}_k = \mathrm{atan2}(v^{j}_{k,y}, v^{j}_{k,x}) \in [0, 2\pi)$ and a speed $\rho^{j}_k = \|\Pi\mathbf{v}^{j}_k\| \in \mathbb{R}^{}_{>0}$, where $\|\cdot\|$ denotes the Euclidean norm. Since the detection times $\{t^j_k\}_{k=1}^{N^j}$ are sparse and irregularly spaced, the flow must be estimated from a limited number of unevenly distributed observations. We take a moving agent as the unit of observation: as stationary agents have no defined heading, they do not contribute to the flow distribution; and, similarly, in-place rotations are outside the scope of the model.

\subsection{Dynamics Field}
\label{sec:dynamics-field}

At any time $t$, each agent $j$ occupies a position $\mathbf{x}^j(t)$ and moves with a velocity $\mathbf{v}^j(t)$. We model this 
as a point process on $\mathbb{R}^3$ whose points are marked by their heading and speed. The quantity to be estimated is the \emph{intensity} of this process at an arbitrary position $\mathbf{x} \in \mathbb{R}^3$, heading $\theta$, speed $\rho$, and time $t$,
\begin{equation}
  \Lambda(\mathbf{x}, \theta, \rho, t) \ge 0 .
  \label{eq:joint-intensity}
\end{equation}
Marginalizing \eqref{eq:joint-intensity} over velocity yields the \emph{occupancy intensity}
\begin{equation}
  \lambda(\mathbf{x}, t) \triangleq \int_0^{2\pi}\!\!\int_0^\infty
    \Lambda(\mathbf{x}, \theta, \rho, t)\, d\rho\, d\theta ,
  \label{eq:presence}
\end{equation}
whose integral over a region is the expected number of agents in that region at time $t$.

Normalizing \eqref{eq:joint-intensity} by \eqref{eq:presence} yields the \emph{flow distribution}
\begin{equation}
  p(\theta, \rho \mid \mathrm{present}, \mathbf{x}, t) \triangleq
    \frac{\Lambda(\mathbf{x}, \theta, \rho, t)}{\lambda(\mathbf{x}, t)} ,
  \label{eq:flow}
\end{equation}
where $\mathrm{present}$ denotes the event that an agent occupies $(\mathbf{x}, t)$ and \eqref{eq:flow} its velocity distribution at that position.
It is a density on the cylinder $S^1 \times \mathbb{R}_{>0}$, normalized jointly over heading and speed by construction 
and it is only defined where $\lambda(\mathbf{x}, t) > 0$.
Equations~\eqref{eq:presence} and~\eqref{eq:flow} are the intensity-mark factorization $\Lambda = \lambda\,p$ of the process, separating \emph{how many} agents pass a location from \emph{how} they move there.




Writing $\mathcal{D}(\mathbf{x}, t)$ for the flow distribution and the occupancy intensity at one position and time, the \emph{dynamics field} $\mathcal{D}$ is distribution-valued with codomain $\left(\mathcal{P}(S^1 \times \mathbb{R}_{>0})\cup\{\varnothing\}\right) \times \mathbb{R}_{\ge 0}$, where $\mathcal{P}(\cdot)$ denotes probability measures on a space and $\varnothing$ represents the positions at which~\eqref{eq:flow} is undefined.

\subsection{Estimation Problem}
\label{sec:estimation-problem}

Given the stream $\mathcal{Z}^{M}_{\le T}$, our goal is to forecast the dynamics field $\mathcal{D}(\mathbf{x}, t)$ at an arbitrary \emph{future} query $(\mathbf{x}, t)$ with $t > T$. 
Note that forecasting beyond the observation horizon 
is well-posed only under recurrence. The assumption is stated on the joint intensity~\eqref{eq:joint-intensity}, a non-negative function of its arguments and therefore additively decomposable. We assume that the intensity $\Lambda$
can be decomposed into a recurrent component $\hat{\Lambda}\ge 0$,
associated with a set of periods fixed from the environment (\eg~daily cycles), and a residual random component $\varepsilon$
without temporal structure:
\begin{equation}\Lambda(\boldsymbol{x},\theta,\rho,t)=\hat{\Lambda}(\boldsymbol{x},\theta,\rho,t)+\varepsilon(\boldsymbol{x},\theta,\rho,t)\,.
\end{equation}
$\hat{\Lambda}$
can be inferred from the observations $\mathcal{Z}^{M}_{\le T}$ and extrapolated to future times, whereas $\varepsilon$
is not forecast. The forecast field $\hat{\mathcal{D}}$ follows from $\hat{\Lambda}$ through~\eqref{eq:presence} and~\eqref{eq:flow}, and carries $\varnothing$ in its flow component wherever the recurrent part places no occupancy. 

Lifelong deployment precludes storing $\mathcal{Z}^{M}_{\le T}$ in full, so any solution takes the form of an \emph{online} estimator: a state $s \in \mathcal{S}$ whose size does not grow with the number of detections, updated recursively as detections $(t^{j}_k, \mathbf{x}^{j}_k, \mathbf{v}^{j}_k)$ arrive
\begin{equation}
  s_k \leftarrow \mathcal{T}\!\bigl(s_{k-1},\, (t^{j}_k, \mathbf{x}^{j}_k, \mathbf{v}^{j}_k)\bigr),
  \label{eq:state-update}
\end{equation}
with $\mathcal{T} : \mathcal{S} \times (\mathbb{R} \times \mathbb{R}^3 \times \mathbb{R}^3) \to \mathcal{S}$, and a readout function $\mathcal{R}$ mapping the state $s$ to the forecast at any future time,
\begin{equation}
  \hat{\mathcal{D}}(\mathbf{x}, t) \triangleq \mathcal{R}\bigl(\mathbf{x}, t,\, s\bigr), \qquad t > T,
  \label{eq:problem}
\end{equation}
with $\mathcal{T}$ and $\mathcal{R}$ to be determined. Memory therefore stays bounded however many detections arrive, and scales instead with the extent of the environment the robot has covered.

For the forecast to inform a robot's decisions, it must capture the direction in which agents move, how that direction changes through time, and where on the robot's own map the estimate applies. The first two are fixed by the field, and the third one depends on the map built.
The distribution-valued codomain $\mathcal{P}(S^1 \times \mathbb{R}_{>0})$ makes the solution \emph{directional}: the forecast is the full, typically multi-modal distribution over heading and speed, and a scalar summary does not lie in that codomain. The future query $t > T$ makes it \emph{time-conditioned}: the distribution is evaluated at $t$, and a long-run average is constant in $t$.

\subsection{Grounding}
\label{sec:grounding}

Let $\mathcal{G}^T$ denote the metric-semantic representation estimated by the robot up to time $T$, $\mathcal{M}^T \subset \mathbb{R}^3$ the region of the environment it has covered, and $\mathcal{A}^T$ the index set of the metric portion of $\mathcal{G}^T$.
The mapping $\Xi^T:\mathcal{M}^T\to\mathcal{A}^T$ carries a covered position to the element of that portion containing it.
An estimate is \emph{grounded} when it is coindexed with $\mathcal{G}^T$ and remains so as the map is updated (\eg~after a loop closure).
\emph{Coindexing} requires the readout~\eqref{eq:problem} to factor as
\begin{equation}
    \mathcal{R}(\boldsymbol{x},t,s)=\tilde{\mathcal{R}}(\Xi^T(\boldsymbol{x}),t,s)\,.
\end{equation}
The forecast can be then addressed by an element of $\mathcal{G}^T$ rather than by a coordinate external to it. As $\Xi^T$ is a component of $\mathcal{G}^T$, it is itself an estimate, which will be revised as the reconstruction is.
Correction of accumulated pose drift displaces geometry already built.
We model it as a warp $\mathcal{W}: \mathbb{R}^3 \to \mathbb{R}^3$, a diffeomorphism acting on $\mathcal{Z}^{M}_{\le T}$ by
\begin{equation}
  \mathcal{W} \cdot \mathcal{Z}^{M}_{\le T}
  \triangleq
  \bigl\{\bigl(t^{}_k,\, \mathcal{W}(\mathbf{x}^{}_k),\,
    \boldsymbol{J}^{}_{\mathcal{W}}(\mathbf{x}^{}_k)\,\mathbf{v}^{}_k\bigr)\bigr\}_{k},
  \label{eq:warp-action}
\end{equation}
displacing positions and transporting velocities by the Jacobian $\boldsymbol{J}^{}_{\mathcal{W}}$, so a correction that rotates a region of the map rotates the observed headings by the same amount. Let $\mathcal{W}^{}_{\!*}$ denote the
pushforward of~\eqref{eq:warp-action} onto the codomain of $\mathcal{D}$, the warp acting at the level of densities.
The estimate is \emph{warp-consistent} when
\begin{equation}
  \mathcal{F}\bigl(\mathcal{W}(\mathbf{x}), t; \mathcal{W} \cdot \mathcal{Z}^{M}_{\le T}\bigr)
  =
  \mathcal{W}^{}_{\!*} \mathcal{F}\bigl(\mathbf{x},\, t; \mathcal{Z}^{M}_{\le T}\bigr),
  \quad \forall\, \mathbf{x} \in \mathcal{M}^T,
  \label{eq:warp-consistency}
\end{equation}
where $\mathcal{F}$ is the composite of the update \eqref{eq:state-update} and the readout \eqref{eq:problem}. Under \eqref{eq:warp-consistency} a correction relocates where a forecast applies and rotates the directions it predicts, 
leaving
the accumulated statistics 
otherwise unchanged.
A correction may carry two elements of $\mathcal{A}^T$ onto one (\eg~after a loop closure that aligns two
observations of the same region), so
$\Xi^T \circ \mathcal{W}$ need not be injective even where $\mathcal{W}$ is.
Equality in~\eqref{eq:warp-consistency} then holds at every corrected index reached from exactly one element; where several
are carried onto one, their accumulated statistics are pooled and~\eqref{eq:warp-consistency} holds for the pooled estimate.
\section{Scene Graph Substrate}
\label{sec:scene-graph}

\Cref{sec:grounding} requires an estimate coindexed with $\mathcal{G}^{T}$ and warp-consistent under corrections applied to it. We take $\mathcal{G}^{T}$ to be a hierarchical 3DSG built online from the robot's sensor stream~\cite{hughes2024foundations}, which supplies an index set, a backend that emits the corrections $\mathcal{W}$, and a navigational layer.

\subsection{Hierarchical 3D Scene Graph}
\label{sec:substrate}

A 3DSG $\mathcal{G}^t \triangleq (\mathcal{V}^t, \mathcal{E}^t)$ organizes the environment into nodes $\mathcal{V}^t = \bigsqcup_{\ell} \mathcal{V}^t_\ell$ at increasing levels of abstraction. Inter-layer edges connect each node to its parent one level above; intra-layer edges relate nodes within a level. The choice of layers is task dependent but can be generalized as a metric substrate, a motion graph, a navigational and action affordance abstraction layer, a semantic abstraction and a global structure~\cite{catalano20253d}. The graph is built incrementally from the sensor stream, so the nodes, the edges, and the geometry beneath them all change as the robot operates. The dynamics field of \Cref{sec:problem} is defined over four of its elements.

\subsubsection{Voxels} At the base of the graph is a Truncated Signed Distance Field (TSDF), which provides the 3D scene geometry from which the higher layers are derived. It partitions the environment into disjoint cells of side $s^{}_v$ indexed by $\mathbf{i} \triangleq (i^{}_1, i^{}_2, i^{}_3) \in \mathbb{Z}^3$, of which the \emph{active voxels} $\mathcal{A}^t$ are those the TSDF has observed up to time $t$.
The set therefore realizes the index set of~\Cref{sec:grounding}, the covered region is the union of its cells $\mathcal{M}^t = \bigcup^{}_{\mathbf{i} \in \mathcal{A}^t} \mathrm{cell}(\mathbf{i})$, and the assignment of a position to the cell containing it
realizes the index map $\Xi^t$. Both sets grow as the robot explores the environment.

\subsubsection{Navigational Nodes} Each node of the navigational layer $\mathcal{V}^t_{\mathrm{nav}}$ is a collision-free location extracted from the Generalized Voronoi Diagram of the reconstructed free space~\cite{hughes2024foundations}. It carries a \emph{clearance}, the distance to the nearest reconstructed surface, and owns a \emph{support set} of the active voxels nearest to it.

\subsubsection{Navigational Edges} An edge denotes straight-line traversability between two navigational nodes~\cite{hughes2024foundations}. Support sets and edges together induce a \emph{traversability relation} over the active voxels: 
two voxels are related when both fall in one node's support set, or in those of two nodes joined by an edge.

\subsubsection{Room Nodes} A room node groups the navigational nodes enclosed by one region of the environment, giving a level of the hierarchy above the navigational layer.

\subsection{Grounding on the Graph}
\label{sec:grounding-on-graph}

This choice of $\mathcal{G}^T$ satisfies the two conditions of \Cref{sec:grounding}.
Coindexing holds as the readout is evaluated at $\mathbf{i} = \Xi^T(\mathbf{x}) \in \mathcal{A}^T$ and because $\Xi^T$ is the scene graph's own cell assignment. The forecast is addressed by a cell from which the higher-abstraction cells are themselves derived, so lifting an estimate to higher-abstraction nodes is a lookup over the set. Warp-consistency holds because each re-optimization of the pose graph warps the reconstructed geometry and, with it, the active voxels, by the action \eqref{eq:warp-action}. Re-keying the cells through every such correction relocates where a forecast applies, leaving the accumulated statistics otherwise unchanged.

No previous map of dynamics satisfies both: grid-based models~\cite{molina2021robotic} keep their indices fixed while the corrected geometry moves away from them, and node-proximity models~\cite{catalano2025aion, catalano2025rheos} re-aggregate observations under a changed navigational layer rather than transporting the estimate (\Cref{sec:related_work}, \Cref{tab:comparison}).

\section{Modeling Spatio-Temporal Flow Dynamics}
\label{sec:dynamics-modelling}

Evaluating the readout~\eqref{eq:problem} at an arbitrary future query is intractable in general: the dynamics field is distribution-valued at every position and every future time, and its joint posterior from sparse, irregular observations is high-dimensional and coupled across space. Kairos makes it estimable through four reductions.
First, the field is piecewise constant on the cells of the reconstruction and estimated independently at each one. Second, at each cell the occupancy intensity~\eqref{eq:presence} separates from the flow distribution~\eqref{eq:flow}, each carried by a \emph{separate} channel.
Third, the flow distribution factorizes over a fixed set of cardinal directions, reducing~\eqref{eq:flow} to one time-varying mixing weight per direction, while the speed along each direction is summarized by a running mean and held time-invariant. 
Fourth, every resulting quantity, the occupancy rate and each directional weight alike, is a scalar trajectory in time, forecast by a single spectral primitive. 

This reduces future-time estimation in~\eqref{eq:problem} to a constant-memory online update that is \emph{directional}, as the per-direction channels retain the full distribution; \emph{time-conditioned}, as the spectral primitive forecasts each channel forward; and \emph{grounded}, as every channel is keyed to the map's cells.
Two further terms act across cells: evidence sharing, which shrinks a sparsely observed cell's weights toward its neighborhood's, and a pairwise dependence relating the flow at adjacent cells.

\subsection{Piecewise-Constant Field}
\label{sec:discretization}

The state factorizes into one fixed-size per-cell state $s^{}_{\mathbf{i}}$, realizing the abstract $s$ of \eqref{eq:state-update}.
A query at $(\mathbf{i}, t)$ retrieves that cell's state and evaluates it at $t$.

The occupancy intensity \eqref{eq:presence} is thus realized per voxel. Piecewise constancy holds $\lambda(\mathbf{x}, t) = \lambda^{}_{\mathbf{i}}(t)$ for all $\mathbf{x}$ in the cell $C^{}_{\mathbf{i}}$, so its integral over a cell collapses to $\int_{C_{\mathbf{i}}} \lambda(\mathbf{x},t),d\mathbf{x} = \lambda_{\mathbf{i}}(t),s_v^3$, the expected number of agents in a voxel of side $s^{}_v$ at time $t$. A count over the cell and the intensity are then the same quantity up to the known cell volume, and the intensity, which is a density and thus determined at a point by no finite set of detections, is instead estimated through the cell count.

\subsection{Flow Dynamics Model}
\label{sec:flow_model}

\begin{figure*}[t]
  \centering
  \begin{subfigure}{0.3\textwidth}
    \includegraphics[width=\linewidth]{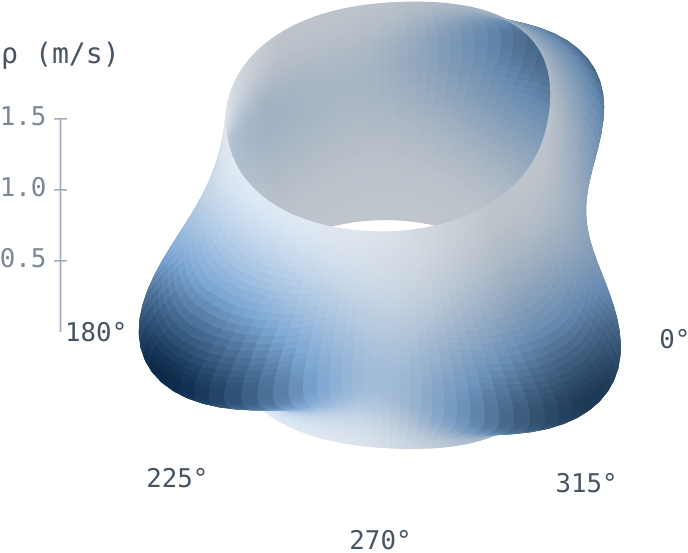}
    \caption{Cylinder}
  \end{subfigure}
  \begin{subfigure}{0.68\textwidth}
    \includegraphics[width=\linewidth]{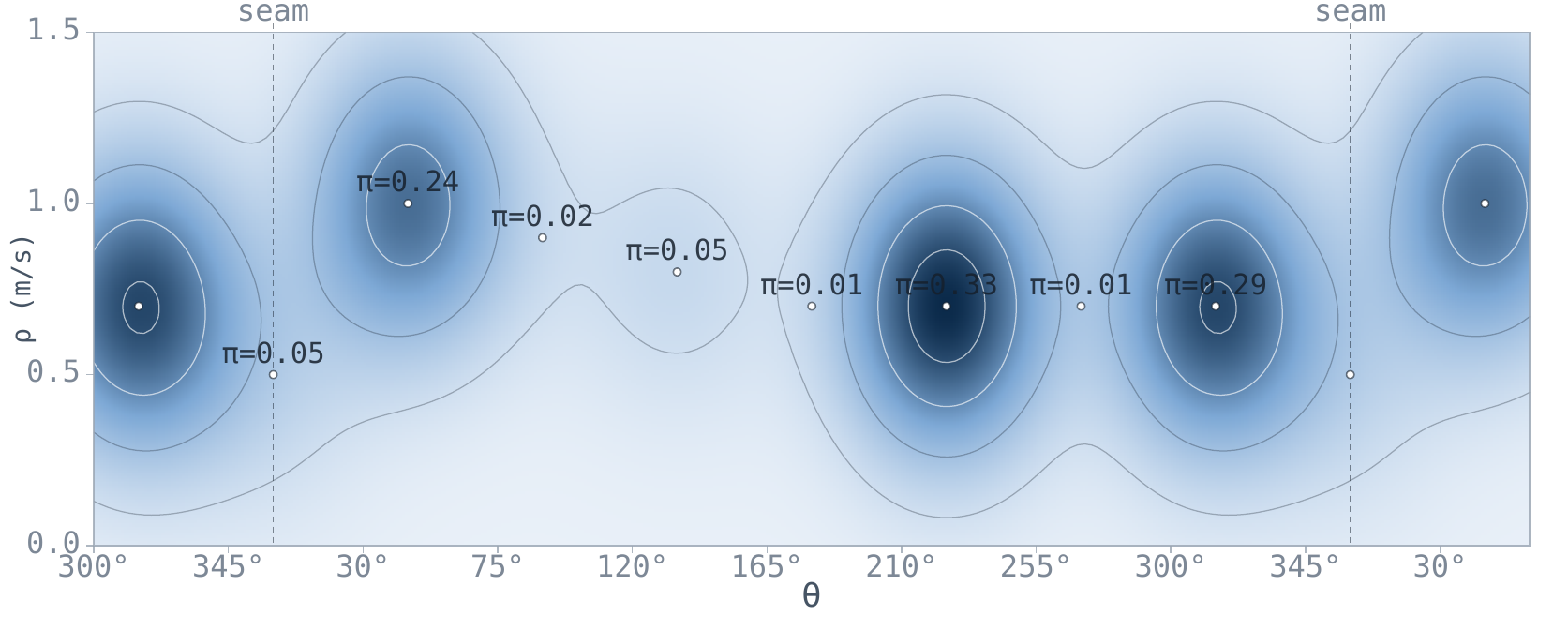}
    \caption{Flat}
  \end{subfigure}
  \caption{Fixed-component SW-GMM for a single voxel, on two embeddings of its domain $[0,2\pi) \times \mathbb{R}^{}_{>0}$, a heading circle times a speed half-line. The $K{=}8$ components sit at cardinal headings $(k-1)\pi/4$ with shared spreads $\sigma^{}_\theta = 0.4$\,rad and $\sigma^{}_\rho = 0.3$\,m/s, so a component varies across voxels only in mean speed $\mu^{(k)}_{\rho,\mathbf{i}}$, and across slots only in that mean and the weight $\pi^{(k)}_{\mathbf{i}}$. 
  Both panels render the same mixture.
  }
  \label{fig:fixed_component}
\end{figure*}

Each active voxel $\mathbf{i} \in \mathcal{A}^t$ allocates, on its first detection, a SW-GMM over the polar decomposition of an agent's velocity vector $(\theta, \rho) \in [0, 2\pi) \times \mathbb{R}^{}_{>0}$ (its  heading and speed respectively). The per-voxel flow distribution \eqref{eq:flow} is:
\begin{equation}
  p^{}_{\mathbf{i}}(\theta, \rho \mid t) \triangleq \sum_{k=1}^{K} \pi^{(k)}_{\mathbf{i}}(t)\, p^{(k)}_{\mathbf{i}}(\theta, \rho),
  \label{eq:mixture}
\end{equation}
with time-varying mixing weights $\pi^{(k)}_{\mathbf{i}}(t) \ge 0$ that form a distribution at every time, $\sum_{k=1}^{K} \pi^{(k)}_{\mathbf{i}}(t) = 1$; the component densities $p^{(k)}_{\mathbf{i}}$ are time-invariant, so only the weights carry the time dependence.
As $\theta$ is circular, each component $k$ is a wrapped bivariate Gaussian, summed over angular windings:
\begin{equation}
  p^{(k)}_{\mathbf{i}}(\theta, \rho) \triangleq
  \sum_{w=-W}^{W}
  \frac{1}{2\pi\sqrt{\det\boldsymbol{\Sigma}}}
  \exp\!\Bigl(-\tfrac{1}{2}
    \mathbf{d}_w^\top \boldsymbol{\Sigma}^{-1} \mathbf{d}_w
  \Bigr),
\end{equation}
where $\mathbf{d}^{}_w = (\theta + 2\pi w - \mu^{(k)}_\theta,\; \rho - \mu^{(k)}_{\rho,\mathbf{i}})^\top$, and we set $W{=}2$. Note that only the angular factor is wrapped (making the mixture semi-wrapped). The radial factor is an ordinary Gaussian on $\rho$, and since walking speeds lie several $\sigma^{}_\rho$ above zero, the mass at $\rho < 0$ is negligible and is left untruncated.

Unlike previous work~\cite{kucner2017enabling, catalano2025rheos}, we fix the number of components $K{=}8$. This makes the component geometry universal: slot $k$ denotes the same heading band in every voxel, so each weight is a scalar signal with a fixed meaning that a single predictor can track through time, and the weight vectors $\pi^{}_{\mathbf{i}}(t)$ are directly comparable across voxels.
Each component $k$ has mean heading $\mu^{(k)}_\theta = (k-1)\pi/4$ ($k = 1,\ldots,K$) and diagonal covariance $\boldsymbol{\Sigma} = \mathrm{diag}(\sigma^2_\theta, \sigma^2_\rho)$, both shared across all voxels, while the mean speed $\mu^{(k)}_{\rho,\mathbf{i}}$ is estimated per voxel: a component depends on the voxel through its radial mean alone (see \Cref{fig:fixed_component}).

We set the angular and radial spreads to $\sigma^{}_\theta = 0.4$\,rad and $\sigma^{}_\rho = 0.3$\,m/s, with zero cross-covariance, and identical across all components and voxels. 
The angular spread is therefore half the $\pi/4$ spacing between adjacent slots, so neighboring components overlap and a heading falling between two cardinals is shared by its two nearest slots. The fixed slots correspond to the orientation histograms of prior temporal maps~\cite{molina2021robotic, catalano2025aion}, realized as overlapping Gaussians rather than disjoint bins.

Heading and speed are not independent in general, and the mixture carries their dependence across slots through the per-slot speed means $\mu^{(k)}_{\rho,\mathbf{i}}$.
A cross-covariance would add dependence within the $\pi/4$ band a single slot spans, where too little heading variation remains for speed to correlate with, so we set it to zero. 
Marginalizing~\eqref{eq:mixture} over speed then leaves a mixture of wrapped normals over heading, and over heading a mixture of Gaussians over speed, under the same weights.
Within this fixed geometry, a voxel's flow is described entirely by the mass each direction holds and the mean speed along it.

\subsection{Temporal Forecasting}
\label{sec:temporal_forecasting}

Because the component geometry is fixed, a voxel's flow distribution \eqref{eq:mixture} changes over time only through its mixing weights. Forecasting the distribution forward in time therefore reduces to forecasting the $K$ mixing-weight trajectories $\pi^{(k)}_{\mathbf{i}}(t)$, one scalar time series per slot. A trajectory is forecastable past the observed interval only where it recurs, so we represent each one in a periodic basis. We adopt the Non-Uniform Discrete Fourier Transform (NUDFT) spectral model~\cite{krajnik2017fremen, molina2021robotic}, instantiating one \emph{predictor} per scalar trajectory; each represents its signal as a constant mean plus harmonics at fixed candidate frequencies.

We fix a set of angular frequencies $\{\omega^{}_f\}_{f=1}^{F}$, with $\omega^{}_f = 2\pi/T^{}_f$ for the periods $T^{}_f$ the environment is expected to exhibit. A predictor ingests timestamped samples $(t^{}_n, y^{}_n)$ of the scalar signal it tracks. For slot $k$ that signal is the fraction of each detection contributing to it: a detection $(\theta^{}_n, \rho^{}_n)$ is assigned softly across the $K$ slots by its \emph{responsibilities}
\begin{equation}
  r^{(k)}_{\mathbf{i},n} \triangleq \frac{p^{(k)}_{\mathbf{i}}(\theta^{}_n, \rho^{}_n)}{\sum_{\ell=1}^{K} p^{(\ell)}_{\mathbf{i}}(\theta^{}_n, \rho^{}_n)},
  \label{eq:responsibility}
\end{equation}
one per slot at the voxel the detection falls in, and slot $k$'s predictor receives one sample per crossing: a traversal $c$ of the voxel contributes a single share $y^{(k)}_c$ (the mean of \eqref{eq:responsibility} over the detections of that traversal) at a time $t^{}_c$ within the visit. The unit of observation is therefore the passage of an agent, so the estimate depends neither on the rate at which detections arrive nor on how long an agent dwells in the cell.

Since the mixing weight is the time-average of these shares over the $C$ crossings observed so far, $\pi^{(k)}_{\mathbf{i}} = \tfrac{1}{C}\sum_{c=1}^{C} y^{(k)}_c$, the predictor's mean term $\gamma^{}_0$ is exactly that running average, the static weight, and its coefficients $\gamma^{}_f$ capture how the share varies in time. Both are updated online as each sample arrives, without accumulating detections into fixed intervals~\cite{molina2021robotic}. Dropping the slot and voxel indices for a generic predictor, and writing  $\gamma_0^-, \gamma_f^-$ for the values before the update:
\begin{equation}
  \gamma^{}_0 \leftarrow \frac{c\,\gamma^-_0 + y_c}{c+1},  \qquad
  \gamma^{}_f \leftarrow \frac{c\,\gamma^-_f +
    (y^{}_c - \gamma^-_0)\,e^{-\mathrm{j}\omega^{}_f t^{}_c}}{c+1},
\end{equation}
where $\mathrm{j}$ is the imaginary unit, and the $\gamma^{}_f$ update accumulates, for each frequency, the correlation between its sinusoid and the signal's deviation $(y^{}_c - \gamma_0^-)$ from that mean.

Prediction at a future time $t$ reconstructs the tracked signal from a \emph{model order} $M \le F$: the $M$ of the $F$ candidate frequencies whose coefficients have the largest magnitude~\cite{krajnik2017fremen}.
The resulting forecast $\hat{y}(t)$ of the tracked signal (here slot $k$'s mixing weight, $\hat{y}(t) = \hat{\pi}^{(k)}_{\mathbf{i}}(t)$) is
\begin{equation}
  \hat{y}(t) \triangleq \gamma^{}_0 +
  \sum_{m=1}^{M} 2|\gamma^{}_{f^{}_m}|\cos\!\bigl(\omega^{}_{f^{}_m} t +
  \angle\gamma^{}_{f^{}_m}\bigr),
  \label{eq:predict}
\end{equation}
where $|\gamma^{}_{f^{}_m}|$ and $\angle\gamma^{}_{f^{}_m}$ are the magnitude and phase of the $m$-th retained coefficient.
A predictor is declared valid once it has accumulated at least $c^{}_{\mathrm{min}}$ samples.
The $K$ slot predictors are fitted independently, so~\eqref{eq:predict} guarantees neither non-negativity nor a unit sum, while~\eqref{eq:mixture} requires the weights to form a distribution.
Therefore, the forecasts $\hat{\pi}^{(k)}_{\mathbf{i}}(t)$ are clipped at zero and re-normalized to the probability simplex before evaluation. The predicted dominant heading $\hat{\theta}^{}_{\mathrm{dom}}(t)$ and speed $\hat{\rho}^{}_{\mathrm{dom}}(t)$ are the weight-averaged component means at the query time.

A voxel also stores the per-slot speed means $\mu_{\rho,\mathbf{i}}^{(k)}$ (\Cref{sec:flow_model}). Unlike the mixing weights, these are not forecast but maintained as the responsibility-weighted running mean of the slot's speeds.
Until a slot has enough observations of its own, it falls back to a global-mean prior: the running mean of speed pooled over all slots and voxels seen so far.

\subsubsection{Model Order Selection}
A fixed global spectral order $M$ fits harmonics to noise wherever a cell's signal is stationary. Rather than validation-based order selection~\cite{krajnik2017fremen, molina2021robotic}, each predictor selects its order online through a predict-then-update \emph{prequential gate}: as each sample arrives, and before it updates the coefficients, the gate scores every candidate order's one-step-ahead prediction of it, so each sample is held out against the state that predicts it. At query time the predictor uses the order with the lowest accumulated one-step error, defaulting to the static mean (order $0$) until a higher order lowers it.

\subsubsection{Predictive Uncertainty}
A predictor returns every forecast with a predictive standard deviation formed from two independent sources: $ \hat{\sigma}(t) \triangleq \sqrt{\sigma^2_{\mathrm{epi}}(t) + \sigma^2_{\mathrm{ale}}}.$
The \emph{epistemic} term $\sigma^2_{\mathrm{epi}}(t)$ is the uncertainty remaining in the predictor's own coefficients, and contracts as samples accumulate. The \emph{aleatoric} term $\sigma^2_{\mathrm{ale}}$ is the spread of the samples about the value the predictor tracks.

At fixed phases $\angle\gamma^{}_f$, the reconstruction \eqref{eq:predict} is linear in the mean term $\gamma^{}_0$ and the harmonic amplitudes $a^{}_f = 2|\gamma^{}_f|$, over the known basis $[1, \cos(\omega^{}_1 t + \angle\gamma^{}_1), \ldots, \cos(\omega^{}_F t + \angle\gamma^{}_F)]$. Under Gaussian observation noise and a Gaussian prior on the coefficients, conjugacy makes the posterior over those $F + 1$ quantities Gaussian and updates it in closed form at each sample~\cite{bishop2006pattern}.
We approximate that posterior by its diagonal, one variance per coefficient, which holds the state and each rank-one update at $O(F)$~\cite{sherman1950adjustment}, and omits the correlations induced by the basis functions not being orthogonal over the sample times. The forecast is linear in these quantities, so its variance follows in closed form:
\begin{equation}
  \sigma_{\mathrm{epi}}^2(t) = \mathrm{var}(\gamma^{}_0) + \sum_{m=1}^{M} \mathrm{var}(a^{}_{f^{}_m})\cos^2(\omega^{}_{f^{}_m} t + \angle\gamma^{}_{f^{}_m}).
  \label{eq:epistemic}
\end{equation}
Each amplitude's variance appears scaled by the square of its basis function at the query time, so a harmonic contributes uncertainty in proportion to its weight in the forecast at that time. The variance state spans all $F$ candidate frequencies while the readout runs over the $M$ retained ones, and for a channel tracked by its mean term alone the sum is empty and the variance reduces to $\mathrm{var}(\gamma_0)$.

Alongside $\gamma^{}_0$ a predictor accumulates the running second moment $m^{}_2$ of the values it receives, so $(m^{}_2 - \gamma_0^2)$ gives their variance in constant memory. Measured on few samples this variance is itself poorly determined, so it is combined with a prior variance $\sigma^2_0$ carrying the weight of $\kappa$ samples of evidence,
\begin{equation}
  \sigma^2_{\mathrm{ale}} \triangleq \frac{\kappa\,\sigma^2_0 + C\,(m^{}_2 - \gamma_0^2)}{\kappa + C},
  \label{eq:aleatoric}
\end{equation}
so a cell reports the prior spread until it has evidence of its own, and its measured spread thereafter, with the prior's weight 
decaying as $\kappa/(\kappa + C)$.

This completes the state of a predictor: the mean term $\gamma^{}_0$, the second moment $m^{}_2$, $F$ complex coefficients alongside their $F+1$ variances, and a sample count. The state is $O(F)$ scalars regardless of how long the stream it consumes runs, which is the constant-memory update \Cref{sec:dynamics-modelling} reduces \eqref{eq:problem} to.

\subsection{Evidence Sharing}
\label{sec:evidence_sharing}

A voxel's mixing weights are estimated from its own crossings alone, and on few crossings that estimate is noisy. As flow varies smoothly over the free space agents traverse, a voxel's neighbors are informative about the same quantity it is estimating, and we treat them as a source of evidence.

We define the neighborhood $\mathcal{N}^{}_{\mathbf{i}}$ by the \emph{traversability} relation of the topological substrate (\Cref{sec:substrate}): a voxel $\mathbf{j}\neq\mathbf{i}$ belongs to $\mathcal{N}^{}_{\mathbf{i}}$ if it lies in the support set of the navigational node $v\in\mathcal{V}_{\mathrm{nav}}$ containing $\mathbf{i}$, or that of a node connected to $v$. The support sets are disjoint and every voxel enters $\mathcal{N}^{}_{\mathbf{i}}$ at most once.
Every such voxel is reachable from $\mathbf{i}$ without crossing a reconstructed surface, so no evidence is borrowed across obstacles.

Let $\gamma_{0,\mathbf{i}}^{(k)}$ be the mean term of voxel $\mathbf{i}$'s predictor for slot $k$, $C^{}_{\mathbf{i}}$ its crossing count (\Cref{sec:temporal_forecasting}), and $\mathcal{N}^{}_{\mathbf{i}}$ its neighborhood as defined above, restricted to the voxels holding at least one crossing of their own. 
The neighborhood estimate pools the neighbors' mean terms for slot $k$, each weighted by the evidence behind it (how many agents have crossed it),
\begin{equation}
    \bar{\gamma}_{0,\mathbf{i}}^{(k)} \triangleq
    \frac{\sum_{\mathbf{j} \in \mathcal{N}^{}_{\mathbf{i}}} C^{}_{\mathbf{j}}\,\gamma_{0,\mathbf{j}}^{(k)}}
         {\sum_{\mathbf{j} \in \mathcal{N}^{}_{\mathbf{i}}} C^{}_{\mathbf{j}}},
    \label{eq:neighbour-estimate}
\end{equation}
the value a single voxel holding the neighborhood's pooled crossings would carry. Where no neighbor holds a crossing, or where the navigational layer does not yet cover $\mathbf{i}$, the neighborhood is empty and the voxel reports its own mean.

Combining the voxel's own mean with the neighborhood estimate, the weight voxel $\mathbf{i}$ reports for slot $k$ is:
\begin{equation}
    \tilde{\gamma}_{0,\mathbf{i}}^{(k)} \triangleq    \frac{C^{}_{\mathbf{i}}\,\gamma_{0,\mathbf{i}}^{(k)} + \nu\,\bar{\gamma}_{0,\mathbf{i}}^{(k)}}
         {C^{}_{\mathbf{i}} + \nu},
    \label{eq:shrinkage}
\end{equation}
where $\nu$ is the weight the neighborhood carries, expressed in crossings and shared across slots and voxels. A voxel with few crossings returns a value close to the neighborhood estimate; one with many reports close to its own.
Both means are distributions over the $K$ slots and \eqref{eq:shrinkage} is a convex combination of them, so the shared weights remain a distribution.

\subsection{Flow Dependence}
\label{sec:flow_dependence}

A voxel's mixture is a \emph{spatially marginal} flow distribution, estimated independently of its neighbors. The sharing of \Cref{sec:evidence_sharing} acts on that marginal, sharpening a voxel's estimate of its own distribution from the evidence around it.
Reasoning jointly over several voxels from marginals alone, however, admits combinations that continuous agent motion precludes (\eg{} an agent heading east at one voxel and west at the next).
Continuous motion instead correlates flow direction across neighboring voxels. Directional and spectral maps of dynamics leave this dependence unmodeled, estimating each cell independently~\cite{krajnik2017fremen, kucner2017enabling, catalano2025rheos, catalano2025aion}; we capture it with a pairwise coupling.

We assume this dependence is local: a voxel's flow is linked only to its immediate neighbors, so two non-adjacent voxels are conditionally independent given the voxels separating them, and couplings are stored only between directly adjacent voxels (\ie~the state grows with the traversable edges rather than with pairs of cells).

As in \Cref{sec:evidence_sharing}, adjacency is \emph{traversable} adjacency, narrowed to face-sharing pairs.
The dependence is induced by one passage crossing both cells in turn: face-sharing supplies that contiguity, while traversability confines the pair to one region of free space.
The coupled field is then a Markov random field over the graph these pairs define, carried on the traversability relation the 3DSG already encodes.

\subsubsection{Joint and Conditional}

Let $k^{}_{\mathbf{i}}, k^{}_{\mathbf{j}} \in \{1,\ldots,K\}$ denote the slots the flow occupies at voxels $\mathbf{i}$ and $\mathbf{j}$. For each slot pair, we introduce a coupling parameter $\phi_{\mathbf{i}\mathbf{j}}^{(k^{}_{\mathbf{i}}, k^{}_{\mathbf{j}})}(t) \in \mathbb{R}$ and model the joint distribution over two adjacent voxels as their independently-estimated mixing weights, reweighted by it:
\begin{equation}\label{eq:coupling}
  p(k^{}_{\mathbf{i}}, k^{}_{\mathbf{j}} \mid t)
  \triangleq
  \frac{1}{Z^{}_{\mathbf{i}\mathbf{j}}(t)}\,
  \pi_{\mathbf{i}}^{(k^{}_{\mathbf{i}})}(t)\, \pi_{\mathbf{j}}^{(k^{}_{\mathbf{j}})}(t)\,
    \exp\!\bigl(\phi_{\mathbf{i}\mathbf{j}}^{(k^{}_{\mathbf{i}}, k^{}_{\mathbf{j}})}(t)\bigr),
\end{equation}
where $Z^{}_{\mathbf{i}\mathbf{j}}(t)$ sums the numerator over the $K^2$ slot pairs ($64$ terms at $K=8$, so evaluated directly).
The coupling acts as a log-bias on the independent product: at $\phi = 0$ the joint reduces to the product of the two weight vectors (independent voxels), positive $\phi$ raises a slot pair's joint probability (co-occurrence, \eg{} aligned headings), and negative $\phi$ lowers it (incompatibility). 
Conditioning on $\mathbf{i}$ occupying slot $k_{\mathbf{i}}$ gives
\begin{equation}\label{eq:conditional}
  p(k^{}_{\mathbf{j}} \mid k^{}_{\mathbf{i}}, t)
  \propto
  \pi_{\mathbf{j}}^{(k^{}_{\mathbf{j}})}(t)\,
  \exp\!\bigl(\phi_{\mathbf{i}\mathbf{j}}^{(k^{}_{\mathbf{i}}, k^{}_{\mathbf{j}})}(t)\bigr),
\end{equation}
that is, $\mathbf{j}$'s own weights are sharpened on the slots that tend to co-occur with $k^{}_{\mathbf{i}}$ and suppressed elsewhere.
The flow density at $\mathbf{j}$ follows as
\begin{equation}\label{eq:conditional2}
    p_{\mathbf{j}}(\theta,\rho\mid k_{\mathbf{i}},t)=\sum_{k_{\mathbf{j}}} p(k_{\mathbf{j}} \mid k_{\mathbf{i}},t) \, p^{(k_{\mathbf{j}})}_{\mathbf{j}}(\theta,\rho \mid t) \,.
\end{equation}
Conditioning is symmetric: fixing $k^{}_{\mathbf{j}}$ instead yields $p(k^{}_{\mathbf{i}} \mid k^{}_{\mathbf{j}}, t)$ by the same construction. Where the marginal mixture \eqref{eq:mixture} gives the flow at a voxel on its own, the conditional \eqref{eq:conditional2} gives the flow at $\mathbf{j}$ once the flow at a neighbor is known.

\subsubsection{Estimating the Coupling}
The coupling is learned online from paired observations on two timescales. The shorter one sets what constitutes a pair: detections in adjacent voxels $\mathbf{i}$ and $\mathbf{j}$ form a pair on edge $(\mathbf{i},\mathbf{j})$ when they fall within a coherence window $\Delta t^{}_{\mathrm{coh}} = 1$\,s of each other, short enough that both plausibly belong to the same passage of flow across the two cells. The longer one sets how often the coupling is measured: pairs are grouped into consecutive \emph{sampling windows} of length $\Delta t^{}_{\phi}$, each yielding one estimate of $\phi$ per slot pair.
Within a window $w$, the outer product of each pair's responsibilities \eqref{eq:responsibility} 
accumulates into a $K\times K$ co-occurrence tensor
\begin{equation}
  \Omega_{\mathbf{i}\mathbf{j}}^{w,(k_{\mathbf{i}}, k_{\mathbf{j}})} \triangleq \sum_{n \in w} r_{\mathbf{i},n}^{(k_{\mathbf{i}})}\, r_{\mathbf{j},n}^{(k_{\mathbf{j}})},
  \label{eq:cooccurrence}
\end{equation}
over the $N_{\mathbf{i}\mathbf{j}}^{w}$ pairs the window holds. Each pair spreads a unit of mass across the tensor ($\sum_{k^{}_{\mathbf{i}},k^{}_{\mathbf{j}}} r_{\mathbf{i},n}^{(k^{}_{\mathbf{i}})} r_{\mathbf{j},n}^{(k^{}_{\mathbf{j}})} = 1$), so $\Omega_{\mathbf{i}\mathbf{j}}^{w}$ is a soft co-occurrence count: the fractional number of pairs in the window in which voxel $\mathbf{i}$ occupied slot $k^{}_{\mathbf{i}}$ while $\mathbf{j}$ occupied $k^{}_{\mathbf{j}}$.
Because $\phi$ is exactly the log-departure from independence in \eqref{eq:coupling}, its natural estimator is the pointwise mutual information~\cite{cover2006elements}, the log-ratio of observed co-occurrence to the count expected under independence~\cite{church1990word}:
\begin{equation}
  \hat{\phi}_{\mathbf{i}\mathbf{j}}^{w,(k^{}_{\mathbf{i}}, k^{}_{\mathbf{j}})}
  \triangleq
  \log \frac{\Omega_{\mathbf{i}\mathbf{j}}^{w,(k^{}_{\mathbf{i}}, k^{}_{\mathbf{j}})} + \epsilon}{\Omega_{\mathbf{i}\mathbf{j}, \mathrm{ind}}^{w,(k^{}_{\mathbf{i}}, k^{}_{\mathbf{j}})} + \epsilon},
  \label{eq:pmi}
\end{equation}
where $\Omega_{\mathbf{i}\mathbf{j}, \mathrm{ind}}^{w,(k^{}_{\mathbf{i}}, k^{}_{\mathbf{j}})} = \bar{\pi}_{\mathbf{i}}^{w,(k^{}_{\mathbf{i}})} \bar{\pi}_{\mathbf{j}}^{w,(k^{}_{\mathbf{j}})}\, N_{\mathbf{i}\mathbf{j}}^{w}$ is the co-occurrence expected if the two slots were independent, with $\bar{\pi}^{w}$ the mean responsibility vectors over that same window and $\epsilon$ a small constant value for smoothing sparse pairs. When a window holds little evidence, both terms vanish together and the smoothing drives $\hat{\phi}^{}_{\mathbf{i}\mathbf{j}}\to 0$, so a barely-observed edge defaults to independence.
Swapping the two voxels transposes both tensors, so $\hat{\phi}_{\mathbf{i}\mathbf{j}}^{(k^{}_{\mathbf{i}}, k^{}_{\mathbf{j}})} = \hat{\phi}_{\mathbf{j}\mathbf{i}}^{(k^{}_{\mathbf{j}}, k^{}_{\mathbf{i}})}$: the edge is undirected, consistent with the symmetric joint \eqref{eq:coupling}.

Each closed window contributes one sample $\hat{\phi}_{\mathbf{i}\mathbf{j}}^{w}$, timestamped at the window's mid-time, to the corresponding NUDFT predictor (\Cref{sec:temporal_forecasting}). Every sample therefore measures the edge's departure from independence over one bounded stretch of time, and the predictor's mean term $\gamma^{}_0$ averages  those measurements while its harmonic terms fit their variation through the day.
To resolve the rhythms being fitted we set the window from the shortest candidate period, $\Delta t^{}_{\phi} = \min_f T^{}_f / S$ with $S = 12$ samples per cycle, tying the sampling rate to $\{T^{}_f\}$.
A window holding a single pair returns $\hat{\phi} = 0$ identically, since its observed co-occurrence and its own marginal product coincide. 
We therefore keep a window that has not reached $N^{}_{\min}$ pairs open past its nominal span until it does, so frequently observed edges sample at the nominal rate while sparse edges lengthen their own windows.

\subsection{Presence Estimation}
\label{sec:presence_estimation}

The mixture of \Cref{sec:flow_model} describes the flow distribution conditional on an agent being present; the occupancy intensity \eqref{eq:presence} of that voxel, and the presence probability derived from it, are estimated separately.

At any fixed location, an agent's motion reduces to three events: entering the voxel that represents the location, remaining for the time taken to cross it, and exiting. Across many agents, a voxel is entered, occupied, and vacated repeatedly, its occupancy set by how often agents enter and how long each stays.
We formulate this motion with queuing theory~\cite{kleinrock1975queueing}: the voxel is the system, each passing agent a customer, and the crossing time (the dwell) the service time.
The superposition of many independent, sparsely visiting trajectories yields Poisson arrivals with time-varying rate $\lambda^{}_a(t)$; crossing times are heterogeneous across agents, so the service-time distribution is left general; and agents traverse a voxel without queuing for access, so the number served concurrently is unbounded, an idealization that costs little since a fine cell holds about one agent at a time and a crowd appears as many occupied voxels. 
These three assumptions specify an $M^{}_t/G/\infty$ queue.

Within the queuing formalism, the occupancy $L$ (the mean number of agents present) is the arrival rate times the mean crossing time, $L = \lambda^{}_a \tau^{}_{\mathrm{cross}}$~\cite{little1961proof}. 
Under a time-varying rate this becomes $\int_0^{\infty} \lambda^{}_a(t - s)\,\bigl(1 - G(s)\bigr)\,ds$, of which the product is the quasi-stationary approximation, exact whenever the rate is constant over one crossing. A voxel is crossed in a fraction of a second, orders of magnitude below the shortest candidate period, so we apply the identity pointwise at each query time. We estimate $L$ and $\tau^{}_{\mathrm{cross}}$ from observations and invert this relation to recover the arrival rate $\lambda^{}_a (t)= L(t)/\tau^{}_{\mathrm{cross}}$.

\begin{figure}[t]
    \centering
    \includegraphics[width=\columnwidth]{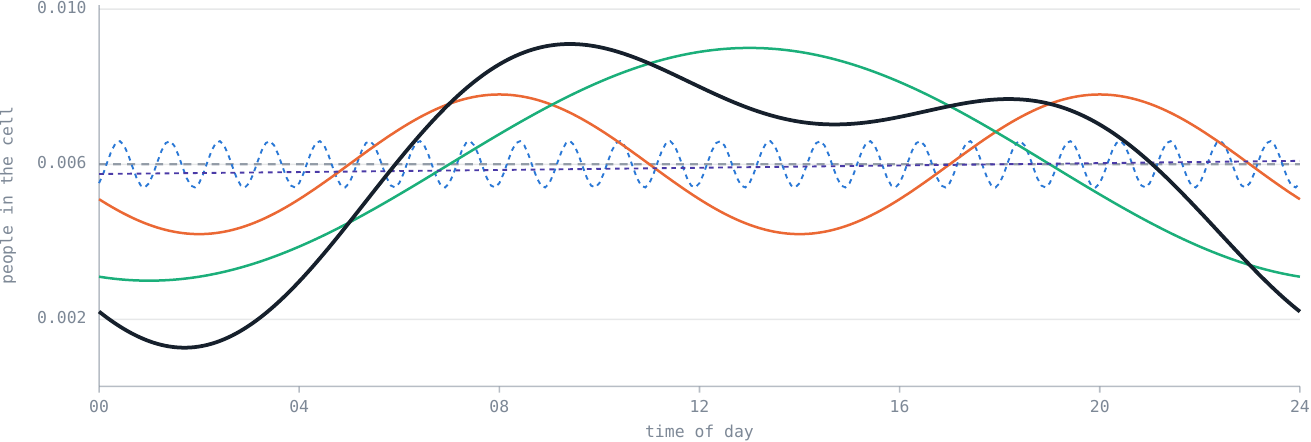}
    \caption{\textbf{Presence channel} in a voxel's state read out across one day, for representative values. Dashed is the stored mean, and the four thin curves are the candidate rhythms of period $1$\,h (blue), $12$\,h (orange), $24$\,h (green) and one week (violet), solid where reported and dashed otherwise; the bold curve is the sum of the reported ones.}
    \label{fig:voxel_forecasting}
\end{figure}

Occupancy is estimated online per voxel with a conjugate Gamma--Poisson model~\cite{jovan2016poisson}: a weakly informative Gamma prior on the detection rate per unit visible time, $\lambda^{}_{\mathrm{vis}} \sim \mathrm{Gamma}(\alpha^{}_0, \beta^{}_0)$ with $\alpha^{}_0 = \beta^{}_0 = 1$. Conjugacy keeps the posterior Gamma, so both parameters update in closed form, $\alpha \leftarrow \alpha + 1$ per detection and $\beta \leftarrow \beta + \Delta t_{\mathrm{vis}}$ per frame in which the voxel is visible, where $\Delta t^{}{\mathrm{vis}}$ is the frame's visible time in seconds, making $\beta^{}_0$ one second of prior exposure.
Multiplying the posterior mode $\hat{\lambda}^{}_{\mathrm{vis}} = (\alpha - 1)/\beta$ by the mean frame duration $\bar{\tau} = (\beta - \beta^{}_0)/n^{}_{\mathrm{vis}}$ gives a mean count per frame, the occupancy $\hat{L}(t) = \hat{\lambda}^{}_{\mathrm{vis}}(t)\,\bar{\tau}$, realizing the occupancy intensity \eqref{eq:presence} over the voxel's extent. 
Once accumulated visible time dominates $\beta^{}_0$, the occupancy is just detections divided by frames in which the voxel was seen, $\hat{L} \approx (\alpha - 1)/n^{}_{\mathrm{vis}}$, empty frames included. We forecast $\hat{\lambda}^{}_{\mathrm{vis}}(t)$ with the spectral predictor (\Cref{sec:temporal_forecasting}), since the static count $(\alpha - 1)/n^{}_{\mathrm{vis}}$ cannot follow a time-varying rate (\Cref{fig:voxel_forecasting}). 

Writing $\bar{\rho}^{}_{\mathbf{i}}$ for the mean speed at voxel $\mathbf{i}$, pooled across slots ---a single time-invariant scalar per voxel--- the crossing time follows from the geometry: an agent crossing a voxel of side $s^{}_v$ at speed $\bar{\rho}^{}_{\mathbf{i}}$ occupies it for $\tau^{}_{\mathrm{cross}} = s^{}_v/\bar{\rho}^{}_{\mathbf{i}}$. The arrival rate is then $\lambda^{}_a(t) = \hat{L}(t)\,\bar{\rho}^{}_{\mathbf{i}}/s^{}_v$.
In an $M^{}_t/G/\infty$ queue the number of agents in the voxel at time $t$ is Poisson with mean $\hat{L}(t)$, and arrivals after $t$ are independent of it.
The probability that at least one agent occupies the voxel at some point in a horizon $\Delta t$ from $t$ therefore has both contributions in one exponent:
\begin{align}\label{eq:voxel-presence}
    p^{}_{\mathrm{present}}(t, \Delta t) 
    &= 1 - \exp\!\left(-\lambda^{}_a(t)\left(\Delta t + \tau^{}_{\mathrm{cross}}\right)\right) \nonumber\\
    &= 1 - \exp\!\Big(-\hat{L}(t)\big(1 + \tfrac{\bar{\rho}^{}_{\mathbf{i}}\,\Delta t}{s^{}_v}\big)\Big).
\end{align}
As $\Delta t \to 0$ this reduces to the instantaneous occupancy $1 - e^{-\hat{L}(t)}$, which is also the value returned when no speed estimate is available and the arrival term drops.

Presence and flow then combine into the full joint distribution as the two factors of a hurdle model, presence multiplying the flow density it conditions:
\begin{equation}
    p(\theta,\rho, \mathrm{present}\mid t, \Delta t) \triangleq   p^{}_{\mathrm{present}}(t,\Delta t)\,p(\theta, \rho \mid  \mathrm{present}, t).
    \label{eq:hurdle}
\end{equation}
This instantiates the readout \eqref{eq:problem} at a cell. The first factor is the presence probability induced over the cell's extent and horizon $\Delta t$ by the occupancy intensity $\lambda$ of \eqref{eq:presence}, realized per voxel as $\hat{L}(t)$; the second is the flow distribution of \eqref{eq:flow}. 
The pair is the dynamics field $\hat{\mathcal{D}}$ at the queried cell and time, returned as a single joint over presence and velocity.

\section{Hierarchical Flow Representation}
\label{sec:hierarchical_representation}

The estimates of \Cref{sec:dynamics-modelling} are held at voxel resolution. We lift them up the layers of \Cref{sec:substrate} so that flow becomes a native attribute of the representation, carried by the navigational layer $\mathcal{V}^t_{\mathrm{nav}}$ and the structure built on it.

\subsection{Graph-Level Flow Dynamics}
\label{sec:lifting}

\subsubsection{Node-Level Aggregation}
Each navigational node aggregates the flow voxels of its support set $\mathcal{S}$ (\Cref{sec:substrate}).
The flow distribution over a region is the intensity-weighted mixture of the distributions of the cells composing it,
\begin{equation}\label{eq:node_mix}
    p^{}_{\mathcal{S}}(\theta,\rho\mid t)
     \triangleq \frac{\sum_{\mathbf{i}\in\mathcal{S}} \hat{L}_{\mathbf{i}}(t)\, p_{\mathbf{i}}(\theta,\rho\mid t)}
         {\hat{L}_{\mathcal{S}}(t)}\,,
\end{equation}
with $\hat{L}_{\mathcal{S}}(t) \triangleq
\sum_{\mathbf{i}\in\mathcal{S}} \hat{L}_{\mathbf{i}}(t)$ the occupancy of the support region. The weights are the cells' occupancies because the joint intensity is additive over the region, $\Lambda^{}_{\mathcal{S}} = \sum_{\mathbf{i}} \lambda^{}_{\mathbf{i}}\,p^{}_{\mathbf{i}}$, and~\eqref{eq:node_mix} is that sum normalized by $\lambda^{}_{\mathcal{S}}$ as in \eqref{eq:flow}. The mixture therefore carries no prediction horizon: how far ahead presence is queried does not enter the distribution over headings.

Given the fixed $K{=}8$ geometry, slot $k$ spans the same heading band in every voxel, and the components differ across voxels only in their radial mean $\mu^{(k)}_{\rho,\mathbf{i}}$ (\Cref{sec:flow_model}). Marginalized over speed, \eqref{eq:node_mix} is therefore a mixture of the same $K$ wrapped normals, and the heading weights reduce exactly to a weighted sum per slot:
\begin{equation}
    \pi^{(k)}_{\mathcal{S}}(t)
    = \frac{\sum_{\mathbf{i}\in\mathcal{S}} \hat{L}_{\mathbf{i}}(t)\, \pi^{(k)}_{\mathbf{i}}(t)} {\hat{L}^{}_{\mathcal{S}}(t)},
    \qquad k = 1,\dots,K \,,
    \label{eq:node_weights}
\end{equation}
so the aggregation is index-aligned. The joint over heading and speed, however, has $K\lvert\mathcal{S}\rvert$ distinct components, one per slot per voxel. We summarize it under the same $K$ slots by aggregating the radial means with the same weights:
\begin{equation}
    \mu^{(k)}_{\rho,\mathcal{S}}(t)
    \triangleq \frac{\sum_{\mathbf{i}\in\mathcal{S}} \hat{L}_{\mathbf{i}}(t)\, \pi^{(k)}_{\mathbf{i}}(t)\, \mu^{(k)}_{\rho,\mathbf{i}}}
           {\sum_{\mathbf{i}\in\mathcal{S}} \hat{L}_{\mathbf{i}}(t)\, \pi^{(k)}_{\mathbf{i}}(t)} \,,
    \label{eq:node_speed}
\end{equation}
the speed mean of slot $k$ pooled over the support region. This is exact in the mean of each slot and approximates its spread.

The presence channel aggregates through the expected count rather than per-voxel. Presence at region scale is
\begin{equation}
p_{\mathrm{present},\mathcal{S}}(t,\Delta t)
  \triangleq 1 - \exp\!\Big(-\hat{L}_{\mathcal{S}}(t)\Big(1 + \tfrac{\bar\rho_{\mathcal{S}}\,\Delta t}{\ell_{\mathcal{S}}}\Big)\Big),
  \label{eq:region_presence}
\end{equation}
with $\ell_{\mathcal{S}}$ the extent of the support region and $\bar\rho_{\mathcal{S}}$ its occupancy-weighted mean speed. 
This reduces to~\eqref{eq:voxel-presence} for a single-voxel support set. It is not the union of the cells' independent presences, which would count an agent once per cell it transits; the region crossing time $\ell_{\mathcal{S}}/\bar\rho_{\mathcal{S}}$ counts that agent once.

These quantities are written into node metadata as the summary tuple $
    \Big(\,\hat{L}^{}_{\mathcal{S}}, \
    \boldsymbol{\pi}^{}_{\mathcal{S}},\
    \theta^{}_{\mathrm{dom}},\ \rho^{}_{\mathrm{dom}},\ c^{}_{\mathrm{dir},\mathcal{S}},\
    t^{}_{\mathrm{latest}}\,\Big)\,$,
where $\theta^{}_{\mathrm{dom}}, \rho^{}_{\mathrm{dom}}$ are the dominant heading and speed, $c^{}_{\mathrm{dir},\mathcal{S}}$ the directional concentration of the region, and $t^{}_{\mathrm{latest}}$ the most recent time the node was observed.
Storing the occupancy $\hat{L}^{}_{\mathcal{S}}$ rather than a presence probability lets a query supply its own $\Delta t$ to \eqref{eq:region_presence} at read time.

The concentration is the occupancy-weighted mean over $\mathcal{S}$, of each voxel's dominant-slot weight $\max_k \pi^{(k)}_{\mathbf{i}}$:
\begin{equation}
    c^{}_{\mathrm{dir},\mathcal{S}}(t)\triangleq \frac{\sum_{\mathbf{i} \in \mathcal{S}} \hat{L}^{}_{\mathbf{i}}(t)\, \max_k \pi^{(k)}_{\mathbf{i}}(t)}
             {\hat{L}^{}_{\mathcal{S}}(t)}.
    \label{eq:concentration}
\end{equation}
It downweights rarely occupied voxels and measures how well-defined the flow is across the node's active region, ranging from $1/K$, where every voxel spreads its mass evenly over the headings, to $1$, where each commits to a single slot.

These per-node attributes form a planner-facing interface: $\hat{L}^{}_{\mathcal{S}}$ and $\theta^{}_{\mathrm{dom}}$ let a route be scored by expected encounters and preferred travel direction, while $\boldsymbol{\pi}^{}_{\mathcal{S}}$ and $c^{}_{\mathrm{dir},\mathcal{S}}$ expose the full directional mix and how sharply voxels commit to it.

\subsubsection{Edge Flow Annotation}

For each intra-layer edge $(u, v)$ whose endpoints carry flow summaries,
forward and reverse flow alignment scores are computed by projecting each slot's heading onto the edge direction:
\begin{equation}
  F^{}_{\mathrm{fwd}}(t) \triangleq \tfrac{1}{2}\sum_{q \in \{u,v\}}
    \sum_k \pi^{(k)}_q(t)\, \hat{L}^{}_q(t)\,
    \max\!\bigl(0,\;\hat{\mathbf{e}}^{}_{uv} \cdot \mathbf{d}^{(k)}\bigr),
\end{equation}
where $\hat{\mathbf{e}}_{uv}$ is the unit edge direction and the direction of slot $k$, $\mathbf{d}^{(k)} = (\cos\mu^{(k)}_\theta, \sin\mu^{(k)}_\theta)^\top$, common to both endpoints since the slot geometry is shared across the map (\Cref{sec:flow_model}).
The reverse score replaces $\hat{\mathbf{e}}_{uv}$ with $-\hat{\mathbf{e}}_{uv}$.
Scaling by the occupancy keeps the score linear in the traffic each endpoint carries, so scores stay comparable across edges.
These augment the navigational edges of \Cref{sec:substrate}, together with average speed and a bidirectionality term.

\subsubsection{Consistency under Optimization}

On every backend optimization the voxel set is re-aligned to the corrected geometry, each cell's index is moved and collisions are pooled. This realizes the positional action of \eqref{eq:warp-action} under any warp.
Because the slot geometry is fixed globally, with slot $k$ denoting the same heading band in every voxel (\Cref{sec:flow_model}), a correction that rotates one region relative to another both displaces that region's cells and turns the headings they store by the same angle.
The rotation is therefore re-expressed on the fixed slots. Each cell takes the rotation the correction applies at its own location from the control points it is deformed against, and its per-slot state is resampled onto the same slots ---by cyclic permutation of the weight vector when the angle is multiple of the $\pi/4$ spacing, and by interpolation between adjacent slots otherwise.
The mixing-weight predictors receive one responsibility vector per crossing at times common to every slot, and their mean and harmonic terms are linear in the value fed, so resampling those coefficients is therefore exact, and returns the same state as accumulating the rotated observations. 
Per-slot speeds are permuted to the nearest slot, since each speed predictor receives only the crossings assigned to its own slot; this leaves at most half the slot spacing of misassignment on that channel. 
Presence carries no direction and is unaffected. 
Both the move and the turn are computed against the cell's original index and frame, so \eqref{eq:warp-consistency} holds wherever the warp is locally a rigid motion whose rotation is a yaw, $\boldsymbol{J}^{}_{\mathcal{W}} \in SO(2)$.
A stored heading is planar, so an out-of-plane component of a correction has no representation on the slots.
Rotating a component of \eqref{eq:mixture} by $\delta$ displaces its mean, so a component left in the frame its observations were accumulated in diverges from the corrected density by $D^{}_{\mathrm{KL}} = \delta^2/2\sigma^2_\theta$, and by convexity the mixture by at most that.

When voxels $\mathbf{i}^{}_a, \mathbf{i}^{}_b$ deform to the same corrected index, they are fused into a single estimate. Every accumulated quantity $x \in \{\gamma^{}_0, \gamma^{}_f, \mu^{(k)}_\rho\}$ is pooled as a sample-weighted average,
\begin{equation}\label{eq:merge-mean}
    \mathrm{pool}(x^{}_a, x^{}_b) \triangleq \frac{N^{}_a\,x^{}_a + N^{}_b\,x^{}_b}{N^{}_a + N^{}_b},
\end{equation}
the value a single voxel of $N^{}_a + N^{}_b$ samples would hold. The count is the one the pooled quantity is accumulated over: the crossings $C$ of \Cref{sec:temporal_forecasting} on the mixing-weight channels, and the detections routed to a slot on that slot's speed channel. The Gamma--Poisson presence counts add, with the shared prior $(\alpha^{}_0, \beta^{}_0)$ subtracted once to avoid double-counting:
$\alpha_{ab} = \alpha_a + \alpha_b - \alpha_0$ and $\beta_{ab} = \beta_a + \beta_b - \beta_0$.
Couplings are addressed by their endpoints' indices, so the same re-keying carries them: an edge follows the corrected indices of the two cells it relates, and edges arriving at one pair pool their co-occurrence~\eqref{eq:cooccurrence} and their coefficients as above.

\subsection{Spatial Motion Semantics}
\label{sec:archetypes}

To expose higher-level crowd structure, per-node flow summaries drive a \emph{motion archetype} layer inserted above the navigational nodes. Each archetype node groups spatially adjacent navigational nodes sharing a motion-semantic class: \emph{Static}, \emph{Unimodal}, \emph{Bimodal}, \emph{Intersection/Multimodal}, or \emph{Diffuse}.

Nodes whose support set holds no observed voxel are excluded.
A node is \emph{Static} if the presence probability~\eqref{eq:region_presence} induced by its occupancy at a fixed classification horizon decreases below $p^{}_{\min}$; since \eqref{eq:region_presence} is monotone in $\hat{L}^{}_{\mathcal{S}}(t)$, this is equivalently a bound on occupancy.
The remaining classes follow from the number of dominant slots $d^{}_{\mathcal{S}}(t) \triangleq \sum_k \mathbf{1}[\pi_{\mathcal{S}}^{(k)}(t) \ge \tau^{}_w]$ with $\mathbf{1}(\cdot)$ the indicator function: \emph{Unimodal}, \emph{Bimodal}, or \emph{Intersection/Multimodal} for $d^{}_{\mathcal{S}}(t) = 1$, $2$, or ${\ge}3$, and \emph{Diffuse} for $d^{}_{\mathcal{S}}(t) = 0$, where the node is occupied but no heading dominates.
The tests are evaluated in order and the classes are disjoint.

\begin{table}[t]
    \centering
    \caption{\textbf{Evaluation datasets.} TBD~\cite{wang2024tbd} is collected by a real moving robot and exercises the embodied pipeline end to end; ATC~\cite{brvsvcic2013person} and 
    HB~\cite{apeltauer2024spatiotemporal} supply the long-horizon recurrence TBD lacks, and are driven through a \emph{grounded simulation}.}
    \label{tab:datasets}
    \scriptsize
    \resizebox{\columnwidth}{!}{
    \begin{tabular}{@{}lcccl@{}}
        \toprule
        Dataset & Extent & Train / Test & Candidate periods & Primary role \\
        \midrule
        TBD
        & 
        ${\sim}213$\,min
        over $3$\,mo. & $3$ / $3$ days & $\{60, 300, 600\}$\,s & flow prediction, cost \\
        ATC
        & $92$ days over ${\sim}1$\,yr & $84$ / $8$ days & $\{1, 12, 24, 168\}$\,h & \makecell[l]{flow, sparse coverage,\\ loop closure, coupling} \\
        HB
        & $11$\,mo.\ continuous & $227$ / $28$ days & $\{1, 12, 24, 168\}$\,h & \makecell[l]{presence, periodicity, coupling} \\
        \bottomrule
        \end{tabular}
    }
\end{table}

\section{Experimental Evaluation}\label{sec:experiments}

We evaluate Kairos along several axes. \Cref{sec:setup} describes the datasets and the online scoring protocol.
We then compare the predicted flow distributions against grid- and graph-based baselines under identical scoring (\Cref{sec:flow_estimation}) and under sparse embodied coverage (\ref{sec:sparse_coverage}), and test whether the learned state survives map corrections by a SLAM backend (\ref{sec:loop_closure_ablation}) ---the property that distinguishes a grounded forecast from one held in a fixed external frame.
We evaluate the presence forecasts against a dedicated spectral occupancy model (\ref{sec:presence_eval}) and close with the computational cost the flow layer adds over the underlying 3DSG (\ref{sec:cost}). 
Finally, \Cref{sec:encounter_planning} applies the representation to a planning task.

\subsection{Datasets and Protocol}
\label{sec:setup}

\Cref{tab:datasets} summarizes the datasets, the train/test split, and the two observability regimes under which all results are reported.

\subsubsection{Datasets}
\label{sec:datasets}

Evaluating Kairos requires a dataset that simultaneously offers (i)~a \emph{moving} sensing platform, (ii)~RGB-D data, (iii)~annotated or recoverable pedestrian trajectories, and (iv)~multi-session temporal structure (repeated visits and time-of-day variation). No single dataset offers all four, so we evaluate on multiple, each covering a different aspect of the framework. 
The TBD pedestrian dataset~\cite{wang2024tbd} exercises the embodied pipeline end to end, from a real moving ZED stereo camera at $10$\,Hz with trajectories lifted directly to 3D ($16$ recordings, ${\sim}213$\,min over six days across three months). It supplies (i)-(iii) and the repeated sessions of (iv) but no recurrence at the periods those sessions resolve, so its results measure the end-to-end feasibility and the relative ordering of methods under identical online scoring.

Long-horizon forecasting is evaluated on ATC~\cite{brvsvcic2013person} and Havl\'i\v{c}k\r{u}v Brod (HB)~\cite{apeltauer2024spatiotemporal} via \emph{grounded simulation}: a simulated robot with bounded field of view is driven over real recorded trajectories and floor maps, reproducing the online, partial-observability setting Kairos targets. The platform patrols a serpentine coverage loop over the navigation graph authored on the floor plans at $1.5$\,m/s, capturing a small fraction of the recorded detections while its short revisit period exposes every cell at every phase of the daily cycle.
ATC is a large shopping mall, its pedestrian trajectories recorded at $10$\,Hz on $92$ days across roughly a year.
HB is a railway-station concourse recorded continuously for eleven months, with full wall-clock timestamps. Its ${\sim}1$\,Hz trajectories are interpolated per track to $10$\,Hz to match the other datasets.

\subsubsection{Evaluation Protocol}
\label{sec:protocol}

On TBD we train on $3$ days from the first two months and test on $3$ held-out days from the final month, so tests are temporally disjoint from training and separated from it by weeks. On ATC we train on $84$ of the $92$ recorded days and test on the $8$ held-out. On HB we train on the $227$ dates with complete camera coverage (Sept. 2022 through Apr. 2023) and test on the $28$ days that follow.
Following established convention~\cite{krajnik2017fremen, molina2021robotic}, candidate periods are set to each dataset's recurrence scale, as shown in~\Cref{tab:datasets}.
 
Within these candidate sets the prequential gate of \Cref{sec:temporal_forecasting} selects a spectral order per cell online. On ATC and HB the weights gate validates the direction harmonics at a median order of $1$ and the presence gate at orders $1$--$2$. On TBD the presence gate holds at order $0$, matching the absence of an occupancy rhythm, while the weights gate concentrates at the maximum order; since the test sessions are recorded weeks later, at which lag harmonics of $60$--$600$\,s carry arbitrary phase, this reflects the training signal rather than forecastable periodicity.
Spatial and angular discretization is common to all methods: every baseline is evaluated at Kairos's $0.4$\,m cell size and the histogram methods at $8$ orientation bins.

\subsubsection{Observability Regimes}
For the grounded simulations we define two regimes.
In the \emph{embodied} regime the platform observes only detections inside its field of view (${\approx}4\%$ of the ATC corpus, ${\approx}7\%$ of HB); in the \emph{full-visibility} regime, where the estimator is trained on every detection. All main results are embodied; full-visibility results appear only where marked, to measure how a mechanism responds to added observability.

\subsection{Flow Prediction}
\label{sec:flow_estimation}

\subsubsection{Baselines}
\label{sec:flow_baselines}

We compare Kairos's flow distributions against four state-of-the-art
methods for representing pedestrian flow, spanning the two spatial substrates Kairos draws on: grid-based and 3D scene graph.
\textbf{CLiFF-map~\cite{kucner2017enabling}}: an offline, batch SW-GMM with no temporal component; a temporally-agnostic SW-GMM reference for directional density estimation.
\textbf{STeF-Map~\cite{molina2021robotic}}: a uniform-grid temporal model with per-cell spectral predictors over orientation histograms; the closest grid-based analogue to our temporal prediction.
\textbf{Aion~\cite{catalano2025aion}}: a 3DSG method attaching discrete per-node orientation histograms with spectral temporal prediction; it isolates a scene-graph substrate carrying a \emph{binned} directional representation rather than a continuous one.
\textbf{Rheos~\cite{catalano2025rheos}}: a 3DSG method using the same SW-GMM as Kairos, fit online by proximity-based association but without temporal forecasting; it isolates the contribution of temporal prediction.

\begin{table*}[t]
    \centering
    \caption{\textbf{Flow prediction} on TBD~\cite{wang2024tbd}, ATC~\cite{brvsvcic2013person} and HB~\cite{apeltauer2024spatiotemporal}. 
    MLPD (nats) per channel,
    circular CRPS of the heading forecast (rad)
    and speed MAE (m/s). 
    \emph{Cov.}\ is 
    the fraction of test detections a method's own representation covers; uncovered detections receive the uniform floor, whose value the \emph{Uniform} row reports.
    Heading-only baselines carry no speed model
    (---); 
    $\times$ marks a method that cannot be fit 
    at a corpus's scale.
    Bold marks the best method per column.}
    \label{tab:real_results}
    \scriptsize
    \setlength{\tabcolsep}{1.5pt}
    \begin{tabular*}{\textwidth}{@{\extracolsep{\fill}}l c ccc cc c ccc cc c ccc cc@{}}
    \toprule
     & \multicolumn{6}{c}{TBD} & \multicolumn{6}{c}{ATC} & \multicolumn{6}{c}{H} \\
    \cmidrule(lr){2-7}\cmidrule(lr){8-13}\cmidrule(lr){14-19}
     & & \multicolumn{3}{c}{MLPD $\uparrow$} & & & & \multicolumn{3}{c}{MLPD $\uparrow$} & & & & \multicolumn{3}{c}{MLPD $\uparrow$} & & \\
    \cmidrule(lr){3-5}\cmidrule(lr){9-11}\cmidrule(lr){15-17}
    Method & Cov.\ $\uparrow$ & H & S & J & CRPS $\downarrow$ & MAE $\downarrow$ & Cov.\ $\uparrow$ & H & S & J & CRPS $\downarrow$ & MAE $\downarrow$ & Cov.\ $\uparrow$ & H & S & J & CRPS $\downarrow$ & MAE $\downarrow$\\
    \midrule
    \emph{Uniform} (reference)                             & --- & $-1.84$ & $-1.10$ & $-2.94$ & $0.79$ & --- & --- & $-1.84$ & $-1.10$ & $-2.94$ & $0.79$ & --- & --- & $-1.84$ & $-1.10$ & $-2.94$ & $0.79$ & --- \\
    \arrayrulecolor{lightgray}
    \midrule
    \arrayrulecolor{black}
    \addlinespace[2pt]
    CLiFF-map$\dagger$~\cite{kucner2017enabling}          & $1.0$ & $-2.18$ & $-2.14$ & $-4.67$ & $\mathbf{0.81}$ & $0.50$ & $\times$ & $\times$ & $\times$ & $\times$ & $\times$ & $\times$ & $\times$ & $\times$ & $\times$ & $\times$ & $\times$ & $\times$ \\
    STeF-Map$\dagger$~\cite{molina2021robotic}            & $1.0$ & $-3.52$ & --- & --- & $0.82$ & --- & $1.0$ & $\mathbf{-1.43}$ & --- & --- & $\mathbf{0.71}$ & --- & $1.0$ & $\mathbf{-1.38}$ & --- & --- & $\mathbf{0.75}$ & --- \\
    Aion$\ddagger$~\cite{catalano2025aion}                 & $0.94$ & $-5.64$ & --- & --- & $0.91$ & --- & $1.0$ & $-1.75$ & --- & --- & $1.02$ & --- & $1.0$ & $-1.75$ & --- & --- & $0.89$ & --- \\
    Rheos$\ddagger$~\cite{catalano2025rheos}           & $0.88$ & $-4.78$ & $-3.08$ & $-7.07$ & $0.88$ & $0.50$ & $\times$ & $\times$ & $\times$ & $\times$ & $\times$ & $\times$ & $\times$ & $\times$ & $\times$ & $\times$ & $\times$ & $\times$ \\
    Kairos$\ddagger$ (ours)                                & $0.93$ & $\mathbf{-1.90}$ & $\mathbf{-0.94}$ & $\mathbf{-2.87}$ & $0.84$ & $\mathbf{0.42}$ & $0.97$ & $-1.48$ & $\mathbf{-0.16}$ & $\mathbf{-1.61}$ & $\mathbf{0.71}$ & $\mathbf{0.19}$ & $0.97$ & $-1.47$ & $\mathbf{-0.65}$ & $\mathbf{-2.10}$ & $0.76$ & $\mathbf{0.33}$ \\
    \bottomrule
    \end{tabular*}
    \par\vspace{2pt}%
    \noindent\parbox{\textwidth}{\scriptsize \textit{Notes}: $\dagger$ indicates \emph{full observability} regime; $\ddagger$ indicates \emph{embodied path} regime.
    }
\end{table*}

\subsubsection{Evaluation Metrics}
\label{sec:flow_metrics}

Kairos predicts a full distribution over flow and presence at future times, so we evaluate with proper scoring rules~\cite{gneiting2007strictly} as they reward calibrated densities and penalize overconfidence. 

\textbf{Mean Log Predictive Density (MLPD), in nats.} The mean log-density a model assigns to flow observations under its predicted distribution. For each held-out observation $(\theta^{}_n, \rho^{}_n)$:
\begin{equation}
  \mathrm{MLPD} \triangleq \frac{1}{N^{}_{\mathrm{test}}}
  \sum_{n=1}^{N^{}_{\mathrm{test}}}
  \log p\!\bigl((\theta^{}_n, \rho^{}_n) \mid \hat{\mathcal{M}}^{}_{\Xi(\mathbf{x}_n)}(t^{}_n)\bigr),
\end{equation}
where $\hat{\mathcal{M}}^{}_{\Xi(\mathbf{x}_n)}(t^{}_n)$ is the predicted mixture at the voxel containing $\mathbf{x}_n$, evaluated at query time $t^{}_n$. We report it per channel ---heading ($p(\theta)$, \emph{H}), speed ($p(\rho)$, \emph{S}), and joint ($p(\theta,\rho)$, \emph{J})--- so heading-only baselines (STeF-Map, Aion) are scored on the first alone and full-distribution methods on all three.

MLPD is computed over \emph{all} held-out detections. A detection at a location unseen during training receives a uniform-density floor $\log p^{}_{\mathrm{floor}} = -\log(2\pi) - \log \rho^{}_{\max}$ over $[0, 2\pi) \times [0, \rho^{}_{\max}]$, with $\rho^{}_{\max} = 3.0$\,m/s, the physical pedestrian speed cap.
The floor is shared across methods so coverage is penalized identically and no method gains by predicting only where it is confident. The heading floor $-\log(2\pi) = -1.84$ and the joint floor $-\log(2\pi\rho^{}_{\max}) = -2.94$ are therefore the score charged for an uncovered detection.
A covered detection is scored under the predicted density, clamped at $10^{-9}$ ($-20.72$ nats), for all methods so that no near-zero density dominates the mean. We report the match rate (fraction of detections covered) alongside MLPD.

\textbf{Circular Continuous Ranked Probability Score (CRPS), in radians.} 
The CRPS scores the entire predicted heading distribution by its arc distance to the observed heading, so mass placed near the  observation improves the score even when the observation itself is not matched exactly. It complements MLPD, which evaluates the density only at the observed heading and cannot distinguish mass placed $30^\circ$ from mass placed $180^\circ$ away. For circular variables~\cite{grimit2006continuous},
\begin{equation}
  \mathrm{CRPS}(F, \theta^{}_n) \triangleq \mathbb{E}^{}_{F}\, d(\Theta, \theta^{}_n) - \tfrac{1}{2}\, \mathbb{E}^{}_{F}\, d(\Theta, \Theta'),
\end{equation}
where $d$ is arc distance on the circle and $\Theta, \Theta' \sim F$ are independent draws from the predicted heading distribution: the expected arc distance of the forecast mass from the observation $\theta^{}_n$, minus a spread term that keeps the score proper. It is distance-aware and bounded by $\pi$, and a uniform forecast scores $\pi/4 = 0.79$\,rad, which is also the floor assigned to uncovered detections, mirroring the MLPD coverage policy.

\textbf{Speed Mean Absolute Error (MAE), in m/s.} The mean absolute error between predicted and observed per-detection speed; a point error is meaningful here, unlike for heading, since per-slot speed is unimodal.
Unlike previous metrics, it is computed on matched detections only, so it carries no coverage penalty and the subsets differ across methods. STeF-Map and Aion carry no speed model and are omitted from this metric.

\subsubsection{Results}
\label{sec:flow_results}

On TBD, Kairos achieves the best MLPD in every channel and the lowest speed MAE (\Cref{tab:real_results}), despite running embodied against baselines given full observability.
Against STeF-Map, Kairos lowers the heading loss from $-3.52$ to $-1.90$, a $1.62$ nat gain corresponding to roughly $5\times$ the density assigned to held-out headings; and its speed MAE of $0.42$\,m/s is $16\%$ below CLiFF-map's $0.50$\,m/s while covering fewer detections ($0.93$ against the grids' $1.0$).
No method improves on the uniform reference of heading ($-1.90$ against $-1.84$) 
or CRPS ($0.81$-$0.91$ against $0.79$): the hall's flow carries no recurring directional structure at the periods $213$ minutes of recording can resolve. Thus, the informative channels for TBD are speed and the joint.

Rheos shares Kairos's SW-GMM representation yet
loses $4.2$ nats on the joint ($-7.07$ against $-2.87$).
The cause is online covariance estimation: a fraction of per-voxel covariances become near-degenerate, concentrating the joint density on a one-dimensional ridge that assigns near-zero probability to off-ridge observations.
The batch fit of CLiFF-map is unaffected, placing the cause in the incremental estimation; Kairos' fixed diagonal components (\Cref{sec:flow_model}) have no off-diagonal term that can degenerate.

On ATC and HB, STeF-Map achieves a slightly better heading MLPD ($-1.43/-1.38$ versus $-1.48/-1.47$).
The gap traces to the fixed-width heading kernel ($\sigma^{}_\theta = 0.4$\,rad), which spreads each training vote and each query density across adjacent slots: Kairos assigns each cell's rarely-used sectors more mass than their empirical share, and correspondingly less mass to the dominant directions that carry most of the traffic, where STeF-Map's histogram concentrates.
On the distance-aware CRPS the two methods are at parity, and the table's marginal gap comes from the uniform floor assigned to Kairos's unmatched detections.
Kairos additionally predicts the heading marginal jointly with calibrated speed and presence densities, neither of which the grid baseline provides.

The training corpora reach $3.3{\times}10^{9}$ (ATC) and $3.2{\times}10^{8}$ (HB). CLiFF-map's per-cell mean-shift and EM buffer the full corpus and scales quadratically in the samples per cell, beyond practical memory at this scale~\cite{kucner2017enabling}; Rheos fits the same SW-GMM incrementally, but scoring the full joint density at ATC's ${\sim}3.4{\times}10^{8}$ test detections requires weeks of compute, and HB's training stream alone exceeds a month. Both therefore appear on TBD only ($\times$ in \Cref{tab:real_results}).

\subsection{Robustness Under Sparse Embodied Coverage}
\label{sec:sparse_coverage}

\begin{figure*}[t]
    \centering
    \begin{subfigure}{0.32\textwidth}
        \includegraphics[width=\linewidth]{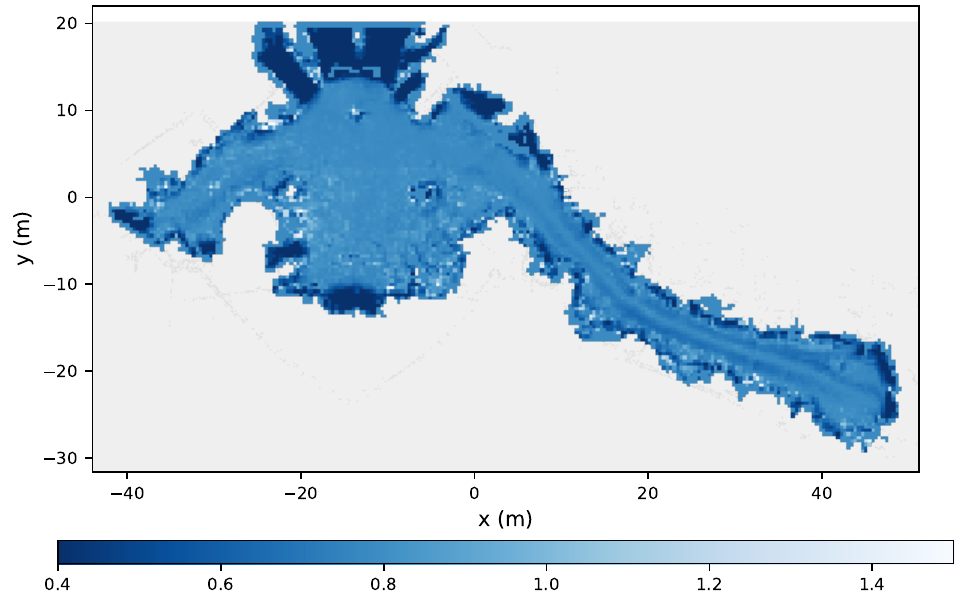}
        \caption{Kairos}
    \end{subfigure}\hfill
    \begin{subfigure}{0.32\textwidth}
        \includegraphics[width=\linewidth]{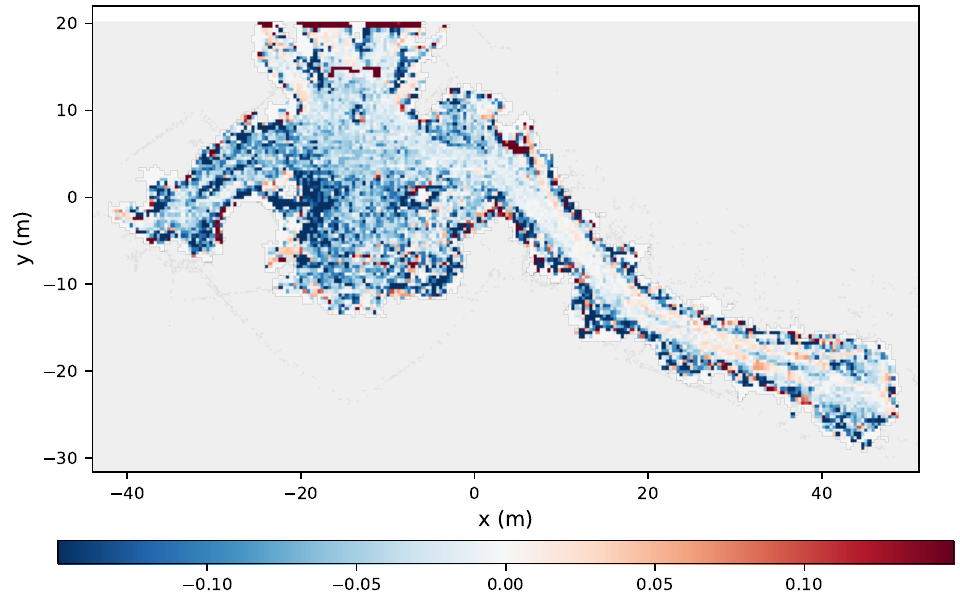}
        \caption{Kairos $-$ STeF-Map~\cite{molina2021robotic}}
    \end{subfigure}\hfill
    \begin{subfigure}{0.32\textwidth}
        \includegraphics[width=\linewidth]{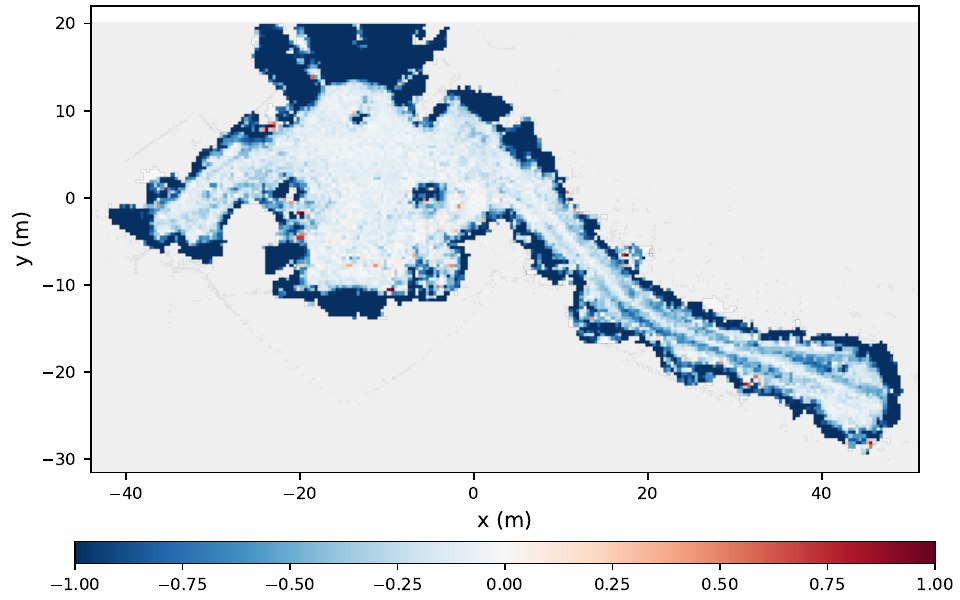}
        \caption{Kairos $-$ Aion~\cite{catalano2025aion}}
    \end{subfigure}
    \caption{\textbf{Per-cell map quality under sparse embodied coverage on ATC~\cite{brvsvcic2013person}.} Circular heading CRPS per cell; gray is unobserved structure. (a) Kairos (ours). (b)--(c) per-cell difference against each baseline.}
    \label{fig:sparse_coverage}
\end{figure*}

The comparison of \Cref{sec:flow_estimation} weights every held-out detection equally, so the few high-traffic cells that every method learns dominate the mean, and Kairos and STeF-Map reach equal CRPS.
This section instead weights every cell equally, in a sparse-mapping regime~\cite{molina2021robotic}, so that rarely-observed cells contribute equally with high-traffic ones. The ATC floor is partitioned by $k$-means into $24$ regions of comparable area and the recorded day into $10$\,min decision intervals. The patrol observes one region per interval and nothing for the remainder of the day, giving an observation budget measured in intervals, set to $4$ of a day's ${\sim}67$ intervals ($6.25\%$). 
At each decision it drives to the chosen region over the traversable graph at $1$\,m/s and covers it with a serpentine loop until the next one.
Over the $84$ training days each region is observed approximately $14$ times: one $10$-minute window every six days. 

Kairos achieves the lowest per-cell heading CRPS (\Cref{tab:sparse_coverage}), $0.72$ against STeF-Map's $0.76$ and Aion's $1.33$.
The gap widens in MLPD: a sparse patrol ($6.25\%$) leaves few detections in a peripheral cell, so most of its $45^\circ$ sectors receive no mass and a held-out heading falling in one of them is scored far below uniform; recorded as $-4.28$ and $-5.21$ against Kairos's $-1.62$, with the uniform reference at $-1.84$. 
Kairos's fixed-width kernel distributes graded mass around each observed heading, retaining a calibrated forecast in cells with few observations. \Cref{fig:sparse_coverage} resolves the comparison spatially: Kairos obtains the lower CRPS in $71\%$ and $96\%$ of cells against STeF-Map and Aion respectively, with the largest margins in the sparsely observed periphery and side passages.

\begin{table}[t]
    \centering
    \caption{\textbf{Map quality under sparse embodied coverage} on ATC~\cite{brvsvcic2013person}. Heading CRPS 
    and heading MLPD (\emph{H}).}
    \label{tab:sparse_coverage}
    \scriptsize
    \setlength{\tabcolsep}{5pt}
    \begin{tabular*}{\columnwidth}{@{\extracolsep{\fill}}lcc@{}} 
        \toprule
        Method & CRPS $\downarrow$ [$95\%$ CI] & H MLPD $\uparrow$ \\
        \midrule
        STeF-Map~\cite{molina2021robotic} & $0.76$ $[0.75, 0.76]$ & $-4.28$ \\
        Aion~\cite{catalano2025aion}      & $1.33$ $[1.32, 1.34]$ & $-5.21$ \\
        Kairos (ours)                     & $\mathbf{0.72}$ $[0.71, 0.72]$ & $\mathbf{-1.62}$ \\
        \bottomrule
    \end{tabular*}
\end{table}

\subsection{Loop Closure Robustness}
\label{sec:loop_closure_ablation}

Odometry drift displaces the map a robot builds, and each loop closure warps it from the previous estimate onto the corrected one. Kairos keys its flow state to the 3DSG, so the flow voxels follow that correction (\Cref{sec:lifting}).

We inject drift into the position and the heading of the pose estimate of the ATC grounded simulation while the platform patrols in the embodied regime of \Cref{sec:setup}. A loop closure is registered at every physical revisit, where corrections are computed from the injected drift and delivered as the time-stamped control-point warp the backend emits. 
The same detection stream, spanning the full $84$ training days, is processed three times: under perfect odometry (\emph{drift-free}), under drift corrected at every loop closure, and under drift left uncorrected within the day.
The uncorrected run is re-anchored only at the start of the next day, as a platform re-localized on start-up would be.

\begin{table}[t]
    \centering
    \caption{\textbf{Loop-closure robustness} on ATC~\cite{brvsvcic2013person}, under a drift model that both displaces and rotates the pose estimate.}
    \label{tab:loop_closure}
    \scriptsize
    \setlength{\tabcolsep}{3pt}
    \begin{tabular*}{\columnwidth}{@{\extracolsep{\fill}}lccccc@{}}
        \toprule
        & \multicolumn{3}{c}{Per detection} & \multicolumn{2}{c}{Per node} \\
        \cmidrule(lr){2-4}\cmidrule(lr){5-6}
        Odometry & H $\uparrow$ & J $\uparrow$ $\downarrow$ & CRPS $\downarrow$ & $\Delta\theta^{}_{\mathrm{dom}}$ ($^\circ$) $\downarrow$ & $>22.5^\circ$ (\%) $\downarrow$ \\
        \midrule
        \emph{Drift-free}    & $-1.47$ & $-1.56$ & $0.71$ & --- & --- \\
        \arrayrulecolor{lightgray}
        \midrule
        \arrayrulecolor{black}
        Uncorrected    & $-1.58$ & $-1.69$ & $0.75$ & $22.8$ & $50.5$ \\
        Loop closure   & $\mathbf{-1.49}$ & $\mathbf{-1.58}$ & $\mathbf{0.72}$ & $\mathbf{9.5}$ & $\mathbf{26.1}$ \\
        \bottomrule
        \end{tabular*}
\end{table}

The corrected run drifts by up to $0.6$\,m ($1.5$ voxels) and $0.77^\circ$ over each $659$\,m patrol loop, and every completed loop triggers a correction. The uncorrected run is assigned a twentieth of each rate, compounding to $\approx3$\,m ($7.5$ voxels) and $4.2^\circ$ by the end of the day.
Both rates are conservative: state-of-the-art odometry drifts by $0.1\%$ to $0.5\%$ of distance traveled and by $0.0013^\circ$ to $0.0033^\circ$ per metre, for LiDAR-inertial and visual-inertial systems respectively~\cite{xu2022fast, zuo2021vins, geiger2012we}, so over the ${\approx}65$\,km the platform covers in a day even the lower rate would displace the map by more than the building's extent.
The corrected run drifts at $0.09\%$ between corrections and the uncorrected run at a twentieth of that, below any published system; its $4.2^\circ$ heading error alone displaces the far side of the $54$\,m floor by a further $4.0$\,m.

Under loop closure the corrected state stays within $0.03$ nats of the reference (\Cref{tab:loop_closure}) in heading MLPD.
Uncorrected, and at a drift rate twenty times lower, the degradation is nearly four times larger at $0.11$ nats, and circular CRPS rises from the reference's $0.71$ to $0.75$.
Each run's state is also aggregated onto the nodes of the navigation graph and compared against the reference's at the nodes both support, through the angle  $\Delta\theta^{}_{\mathrm{dom}}$ between the two dominant directions, reported as its median and as the share of nodes exceeding $22.5^\circ$, half a direction slot. Under loop closure the median is $9.5^\circ$ and a quarter of the nodes exceed the slot; uncorrected, the median is $22.8^\circ$ and half of them do.
Corrections of the form the backend produces therefore preserve the learned state, and the re-keying they require is necessary at any realistic drift rate.

\subsection{Presence Estimation}
\label{sec:presence_eval}

\begin{table*}[t]
    \centering
    \caption{\textbf{Presence estimation} on HB~\cite{apeltauer2024spatiotemporal} across event windows $\Delta t$: MLPP (in nats), {reliability} (Rel), and {resolution} (Res); all in units of ${10^{-3}}$.
    Reliability and resolution are shown from $60$\,s onward, the windows in which they separate the estimators. 
    }
    \label{tab:presence}
    \scriptsize
    \setlength{\tabcolsep}{4pt}
    \begin{tabular*}{\textwidth}{@{\extracolsep{\fill}}cl ccccccccccc@{}}
        \toprule
        & & $\Delta t = 5$\,s & $\Delta t = 10$\,s & \multicolumn{3}{c}{$\Delta t = 60$\,s} & \multicolumn{3}{c}{$\Delta t = 300$\,s} & \multicolumn{3}{c}{$\Delta t = 600$\,s} \\
        \cmidrule(lr){3-3}\cmidrule(lr){4-4}\cmidrule(lr){5-7}\cmidrule(lr){8-10}\cmidrule(lr){11-13}
        Regime & Model & MLPP$\uparrow$ & MLPP$\uparrow$ & MLPP$\uparrow$ & Rel$\downarrow$ & Res$\uparrow$ & MLPP$\uparrow$ & Rel$\downarrow$ & Res$\uparrow$ & MLPP$\uparrow$ & Rel$\downarrow$ & Res$\uparrow$ \\
        \midrule
        \multirow{4}{*}{\makecell[c]{\emph{Embodied}\\\emph{$5$\,m FoV}}}
        & Constant base rate       & $-38$ & $-64$ & $-222$ & $0.1$ & $0.0$ & $-489$ & $0.8$ & $0.0$ & $-601$ & $1.6$ & $0.0$ \\
        & Kairos (no dwell)        & $-45$ & $-85$ & $-421$ & $3.3$ & $0.0$ & $-1421$ & $36.0$ & $0.0$ & $-2153$ & $80.2$ & $0.0$ \\
        & Kairos (static rate)     & $-34$ & $-57$ & $-190$ & $0.1$ & $3.6$ & $-407$ & $3.6$ & $26.1$ & $-508$ & $9.0$ & $44.3$ \\
        & Kairos                   & $\mathbf{-33}$ & $\mathbf{-56}$ & $\mathbf{-182}$ & $0.1$ & $5.1$ & $\mathbf{-379}$ & $1.8$ & $35.5$ & $\mathbf{-460}$ & $4.6$ & $58.9$ \\
        \midrule
        \multirow{5}{*}{\makecell[c]{\emph{Full}\\\emph{visibility}}}
        & Constant base rate       & $-37$ & $-63$ & $-217$ & $0.1$ & $0.0$ & $-481$ & $0.5$ & $0.0$ & $-592$ & $1.0$ & $0.0$ \\
        & FreMEn~\cite{krajnik2017fremen} occupancy$\dagger$ & $\mathbf{-32}$ & $\mathbf{-53}$ & $\mathbf{-172}$ & $0.0$ & $5.6$ & $\mathbf{-343}$ & $0.2$ & $41.5$ & $\mathbf{-393}$ & $0.2$ & $72.5$ \\
        & Kairos (no dwell)        & $-44$ & $-83$ & $-409$ & $3.1$ & $0.0$ & $-1380$ & $34.0$ & $0.0$ & $-2090$ & $75.9$ & $0.0$ \\
        & Kairos (static rate)     & $-33$ & $-56$ & $-185$ & $0.1$ & $3.6$ & $-396$ & $3.8$ & $26.4$ & $-495$ & $9.3$ & $45.2$ \\
        & Kairos                   & $\mathbf{-32}$ & $-54$ & $-175$ & $0.1$ & $5.2$ & $-361$ & $1.9$ & $37.4$ & $-434$ & $4.6$ & $63.9$ \\
        \bottomrule
    \end{tabular*}
    \par\vspace{2pt}%
    \noindent\parbox{\textwidth}{\scriptsize\textit{Notes}: $\dagger$ FreMEn divides each cell's occupied time by 
    total elapsed time, so time in which the cell was not observed is indistinguishable from time in which it was empty. It is therefore fit only 
    under \emph{full visibility}.
    }
\end{table*}

A $0.4$\,m voxel is crossed in a fraction of a second, so instantaneous occupancy is near zero and carries no useful signal and the presence forecast is windowed. 
The target is the event that a cell holds at least one detection within the window $\omega=[t, t+\Delta t]$; we call $\Delta t$ the \emph{horizon} and evaluate $\Delta t \in \{5, 10, 60, 300, 600\}$\,s, the scale at which a navigating robot queries the cells ahead of it. The set is capped at $600$\,s so every window stays below the shortest candidate period on HB ($1$\,h).
Scoring follows the protocol established for spectral occupancy models~\cite{krajnik2017fremen}: the trained state is frozen, and at every step $t$ each covered cell's predicted occupancy probability for $\omega$ is scored against the realized occupancy of that window.

\subsubsection{Baselines}
\label{sec:presence_baselines}

We compare against two baselines and two ablations of our own estimator.
\textbf{Constant base rate}: the training corpus's occupancy frequency at each horizon. Calibrated by construction, it carries no cell-level information.
\textbf{FreMEn~\cite{krajnik2017fremen}}: an independent per-cell spectral occupancy model, the standard temporal-map baseline.
\textbf{Kairos (no dwell)}: the presence estimator without the dwell correction from the static rate, \ie~$\tau_{\mathrm{cross}}=0$, isolating what the correction contributes.
\textbf{Kairos (static rate)}: the presence estimator with the dwell correction but the rate held time-invariant, dropping the gate-selected harmonics and isolating  the periodic component.

\subsubsection{Evaluation Metrics}
\label{sec:presence_metrics}

Presence forecasts are probability forecasts of a binary event, so we score them with a proper scoring rule and decompose that score into its calibration and information components.

\textbf{Mean Log Predictive Probability (MLPP), in nats.} The mean logarithmic score~\cite{gneiting2007strictly} of the realized occupancy of $\omega$ under the predicted probability.
It is maximized only by the true probability and dominated by confident errors on the rare occupied cells.

\textbf{Calibration and Informativeness.} Alongside the MLPP we report \emph{reliability} and \emph{resolution} terms of the vector partition of the Brier score~\cite{murphy1973new}, computed on the same forecast. 
Reliability is the mean squared gap between predicted probabilities and the occupancy frequencies observed at those probabilities ($0$ for a perfectly calibrated forecast); resolution is the spread of those observed frequencies around the overall occupancy rate ($0$ for a constant forecast).
Together they separate whether an estimator is calibrated from whether it is informative. 
We report both from the $60$\,s horizon onward: at $5$ and $10$\,s every estimator's reliability is at most $0.1$ and its resolution at most $0.2$, so neither term separates the estimators there.

\subsubsection{Results}
\label{sec:presence_results}

Presence is evaluated on HB: the occupancy channel carries its strongest temporal structure at the concourse, whose continuous recording contains the full daily cycle. ATC is recorded during opening hours only.
TBD is visited too sparsely to support an occupancy forecast at all, and is retained for the flow metrics.

In the embodied regime Kairos achieves the best MLPP at every horizon (\Cref{tab:presence}), from $-33$ at $5$ to $-460$ at $600$, ahead of its static-rate variant by $1$ and by $48$ respectively.
Kairos is the most informative forecast and the best calibrated: at $600$\,s, its resolution is $58.9$ against the static-rate variant's $44.3$, and its reliability $4.6$ against $9.0$.  
These margins grow with the horizon because a longer window integrates the rate over more of the daily profile, where time-varying and state rates differ most.
The margin is carried by the harmonics the presence gate validates, at a median order of $1$--$2$ on HB (\Cref{sec:protocol}): the daily cycle and its first overtone, the component the static-rate variant holds constant by construction.

The dwell correction is what makes this channel informative. Without it \eqref{eq:voxel-presence} reduces to the instantaneous occupancy $1-e^{-\hat{L}}$, and $\hat{L}\ll1$ at every cell, so forecasts fall in the near-zero bin regardless of query time or cell. Resolution is $0$ at all horizons and MLPP falls to $-2153$, the only estimator below the constant base rate in both regimes.

At full visibility FreMEn achieves the best MLPP at every horizon ($-32$ to $-393$), under two conditions: it is fit on the complete occupancy record of every cell, and the occupancy event is the only measure it models, while Kairos's state additionally yields expected counts and the direction-resolved flow.
Kairos matches it at $5$\,s ($-32$) and falls below it as the window grows, from $-54$ against $-53$ at $10$\,s to $-434$ against $-393$ at $600$\,s. 
The dwell-corrected estimators remain well calibrated: reliability is at most $9.3$ for the Kairos rows and at most $0.2$ for FreMEn's, both far below the resolution attained at the long horizons ($63.9$ and $72.5$ at $600$\,s).

\subsection{Computational Cost}\label{sec:cost}

We profile Kairos following~\cite{hughes2024foundations}, timing every component per frame against a $200$\,ms key-frame budget ($5$\,Hz; the TBD detection stream arrives at $10$\,Hz) on an Intel Core i9-14900KF CPU (\Cref{tab:timing}).
The 3DSG front-end (reconstruction and place extraction) runs at a mean of $76.9$\,ms, inside the budget; the backend (room detection and pose-graph optimization) runs asynchronously and, as in Hydra~\cite{hughes2024foundations}, does not block it.
The flow layer adds $0.43$\,ms per frame, two orders of magnitude below the front-end and $4\,\mu$s per detection.
Node aggregation ($11.6$\,ms) and the deformation remap ($0.3$\,ms) run per backend event against its mean $292$\,ms, adding $4.1\%$ per re-optimization.
Enabling the layer leaves the front-end timing, tail included, unchanged.

\subsection{Downstream Task: Encounter-probability Planning}
\label{sec:encounter_planning}

\begin{table}[t]
    \centering
    \caption{\textbf{Computational cost} on TBD~\cite{wang2024tbd} over all 16 scenes.
    Kairos's \emph{per-frame} row is its full per-frame cost; the remaining
    rows are unit costs and do not sum to it.}
    \label{tab:timing}
    \scriptsize
    \setlength{\tabcolsep}{2.5pt}
    \begin{tabular*}{\columnwidth}{@{\extracolsep{\fill}}cl ccc@{}}
                \toprule
                & & mean$\pm$std [ms] & med [ms] & p95 [ms] \\
                \midrule
    \multirow{2}{*}{\makecell[c]{\emph{Base}\\\emph{3DSG}}}
                & Frontend total & $76.9\pm45.1$ & $74.7$ & $263.2$ \\
                & Backend optimization & $292.0\pm158.2$ & $263.8$ & $989.8$ \\
                \midrule
    \multirow{6}{*}{\makecell[c]{\emph{Flow}\\\emph{dynamics}}}
                & Per frame (10\,Hz) & $0.427\pm0.181$ & $0.441$ & $0.689$ \\
                & Cell update (per det.) & $0.004\pm0.004$ & $0.004$ & $0.010$ \\
                & Sharing (per neighborhood) & $0.001\pm0.001$ & $0.000$ & $0.003$ \\
                & Coupling (per frame it runs) & $0.007\pm0.007$ & $0.006$ & $0.019$ \\
                & Node summary (per event) & $11.596\pm4.727$ & $12.540$ & $16.314$ \\
                & Deformation remap (per event) & $0.328\pm3.079$ & $0.175$ & $0.230$ \\
                \bottomrule
            \end{tabular*}
\end{table}

We evaluate Kairos on the encounter-probability planning task of~\cite{tipaldi2011want}. Starting from a home position at a given departure time, the robot has a $600$ s deadline to reach within $2$ m of a person. A finite-horizon dynamic program decides at each step whether to wait or move to an adjacent cell of the traversable graph; the reward for occupying a cell is the information source's predicted activity there at that time, and the policy maximizes accumulated reward within the deadline.

\subsubsection{Baselines}
The information sources cover the alternatives a robot could carry.
\textbf{None (wait)} and \textbf{None (random walk)} use no information and serve as uninformed baselines.
\textbf{Static rate} uses each cell's training-mean detection rate: where people are on average, with no notion of time.
\textbf{FreMEn}~\cite{krajnik2017fremen} under its full-visibility fit.

\subsubsection{Evaluation Protocol}
We follow the protocol of~\cite{krajnik2015waldo}: a single fixed planner, with only the information source varying across variants. Whereas prior evaluations~\cite{tipaldi2011want, krajnik2015waldo} adopted a simulator with scripted agents and on binary room-occupancy records, we replay every episode against the recorded pedestrians of a held-out day. Episodes start every $30$ minutes within operating hours on each held-out day from a single fixed home node, giving $952$ episodes over $28$ held-out days on HB and $192$ over $8$ days on ATC. For each episode we record whether a person is reached before the deadline (Success), the time to the first encounter (if successful, Med.\ $t$), and the number of distinct people encountered (Enc.). Only the presence channel is evaluated, since an encounter depends on presence alone.

The informed sources drive a deterministic planner over a frozen trained state, so each episode's outcome is fixed.
Because every day repeats the same grid of departure times, the held-out day rather than the individual episode is the unit of replication: we report each success rate with a $95\%$ confidence interval formed as a $t$-interval over the $D$ held-out days. These intervals therefore quantify day-to-day generalization; the single fixed home node means they do not capture sensitivity to the start location. Median time-to-encounter is a discrete statistic and is reported as a point value.

\newcommand{\sci}[3]{\makecell[tc]{$#1$\\[1pt]{\scriptsize$[#2,\,#3]$}}}
\begin{table}[t]
    \centering
    \caption{\textbf{Encounter-probability planning} on HB~\cite{apeltauer2024spatiotemporal} and ATC~\cite{brvsvcic2013person}: success rate,
    median time to the first encounter
    and mean number of distinct people encountered per episode.
    }
    \label{tab:people_finding}
    \scriptsize
    \setlength{\tabcolsep}{2.5pt}
    \begin{tabular*}{\columnwidth}{@{\extracolsep{\fill}}lcccccc@{}}
    \toprule
     & \multicolumn{3}{c}{HB} & \multicolumn{3}{c}{ATC} \\
    \cmidrule(lr){2-4}\cmidrule(lr){5-7}
    Information source & Succ.\ $\uparrow$ & Med.\ $t$ [s] $\downarrow$ & Enc.\ $\uparrow$ & Succ.\ $\uparrow$ & Med.\ $t$ [s] $\downarrow$ & Enc.\ $\uparrow$ \\
    \midrule
    None (wait)             & $0.484$ & $138.0$ & $1.5$  & $0.818$ & $111.0$ & $3.2$ \\
    None (random walk)      & $0.893$ & $72.0$  & $9.2$  & $0.875$ & $45.2$  & $37.9$ \\
    Static rate             & $0.909$ & $\mathbf{18.5}$ & $23.5$ & $0.885$ & $\mathbf{6.5}$ & $14.3$ \\
    FreMEn                  & $\mathbf{0.937}$ & $30.0$ & $\mathbf{27.4}$ & $\mathbf{0.896}$ & $7.0$ & $\mathbf{152.9}$ \\
    Kairos                  & $0.922$ & $30.0$ & $26.8$ & $0.891$ & $7.0$ & $53.0$ \\
    \bottomrule
    \end{tabular*}
\end{table}

\subsubsection{Results}

On HB (\Cref{tab:people_finding}), the static rate succeeds on $91\%$ of episodes and encounters $23.5$ people, against $89\%$ and $9.2$ for the random walk and $48\%$ and $1.5$ for waiting at the home position, showing that knowing where people are on average raises the number of people met rather than the probability of meeting any.
Conditioning on time of day improves both: FreMEn succeeds on $94\%$ of episodes ($95\%$ CI $[0.92, 0.95]$ over the $28$ test days) and Kairos on $92\%$ ($[0.91, 0.94]$), their intervals overlapping, meeting $27.4$ and $26.8$ people ($17\%$ and $14\%$ above the static rate).
The static source reaches its first person sooner ($18.5$\,s median against $30$\,s) but is the least consistent across days: its interval is much wider ($[0.85, 0.97]$) and its per-day success ranges from $0.18$ to $0.97$, because its single greatest-average-rate cell is fixed (failing on days that cell is quiet) whereas the time-conditioned sources relocate by hour and hold near $0.9$ on every day. Kairos remains within $1.6\%$ and $0.6$ encounters of FreMEn, which is fit on the full detection record while Kairos observes ${\approx}7\%$ of the stream on HB and ${\approx}4\%$ on ATC.

On ATC all informed sources succeed nearly identically ($88\%$--$90\%$, with fully overlapping $95\%$ intervals: $[0.86, 0.91]$ for the static rate and $[0.87, 0.92]$ for Kairos over the $8$ test days) and reach the first person within seconds ($6.5$--$7.0$\,s median). ATC's within-day structure is close to spatially uniform, so conditioning on time does not change where the robot should stand: at this pedestrian density place alone suffices. 
Distinct-people counts vary widely across sources ($3.2$ for waiting to $152.9$ for FreMEn), reflecting movement and how often d plan relocates through a dense crowd rather than the quality of its information.
\section{Ablation Studies}\label{sec:ablations}

\begin{table*}[t]
    \centering
    \caption{\textbf{Temporal forecasting of the direction mixing weights} on HB~\cite{apeltauer2024spatiotemporal}: change in joint MLPD from replacing static weights (order-0) with the gated NUDFT forecast (order-1), by hour of day. {Entries are in units of} ${10^{-3}}$ {nats}; positive favors the forecast. \emph{All} is the whole test set over the full 24-hour day.}
    \label{tab:temporal}
    \scriptsize
    \setlength{\tabcolsep}{3pt}
    \begin{tabular*}{\textwidth}{@{\extracolsep{\fill}}l*{20}{c}@{}}
        \toprule
        Regime & 5 & 6 & 7 & 8 & 9 & 10 & 11 & 12 & 13 & 14 & 15 & 16 & 17 & 18 & 19 & 20 & 21 & 22 & 23 & All \\
        \midrule
        Embodied ($5$\,m FoV) & $21.4$ & $15.4$ & $1.0$ & $-1.2$ & $12.6$ & $9.8$ & $13.8$ & $11.0$ & $21.9$ & $13.2$ & $11.7$ & $12.1$ & $10.4$ & $16.2$ & $18.0$ & $-1.6$ & $14.8$ & $-6.2$ & $23.8$ & $\mathbf{11.7}$ \\
        Full visibility & $24.1$ & $17.8$ & $1.3$ & $-0.7$ & $14.3$ & $11.5$ & $16.3$ & $13.8$ & $25.5$ & $14.8$ & $14.2$ & $14.5$ & $13.4$ & $19.5$ & $21.1$ & $0.2$ & $17.7$ & $-5.7$ & $24.4$ & $\mathbf{13.9}$ \\
        \bottomrule
    \end{tabular*}
\end{table*}

\subsection{Temporal forecasting of the direction weights}
\label{sec:direction_forecasting}

This ablation isolates the spectral forecasting of the direction mixing weights against an order-0 arm holding them at their running mean, on HB (\Cref{tab:temporal}).
We report hourly columns between 05--23\,h, since overnight hours carry negligible traffic. The embodied regime is the operating condition; full-visibility regime measures how the same mechanism responds to added observability.

The spectral predictor improves the joint MLPD in both regimes, and the margin grows with observability: by $0.0117$ embodied ($-2.0770$ against $-2.0887$) and $0.0139$ at full visibility ($-2.0728$ against $-2.0867$).
Across the day the improvement is positive at all but three hours (marginal dips at 08:00, 20:00, and 22:00, none beyond $0.007$; \Cref{tab:temporal}), and largest at 05:00, 13:00 and 23:00 (up to $+0.024$ embodied, $+0.026$ at full visibility).
The spectral arm converges to the static weights where the rhythm is absent, so the harmonic terms add resolution where the data supports them and leave the forecast unchanged elsewhere.

A model-free measurement of the direction rhythm on the training detections, the traffic-weighted divergence between each hour's heading histogram and the cell's all-hour marginal, places the recoverable direction-mix signal at $0.011$ bits ($0.008$ nats) on HB and $0.0099$ bits on ATC, which are the cells a detection-weighted MLPD average reports. The $0.0117$ and $0.0139$ nats recovered here (last column in~\Cref{tab:temporal}) therefore sit at the upper end of the available signal.

\begin{table*}[t]
    \centering
    \caption{\textbf{Evidence sharing ablation} on TBD~\cite{wang2024tbd}, ATC~\cite{brvsvcic2013person}, and HB~\cite{apeltauer2024spatiotemporal}: joint MLPD by per-voxel coverage and global MLPD. Coverage strata are dataset-specific, reflecting the corpora's different densities. 
    }
    \label{tab:evidence_sharing}
    \scriptsize
    \setlength{\tabcolsep}{3pt}
    \begin{tabular*}{\textwidth}{@{\extracolsep{\fill}}l cccc cccc cccc@{}}
        \toprule
         & \multicolumn{4}{c}{TBD} & \multicolumn{4}{c}{ATC} & \multicolumn{4}{c}{HB} \\
        \cmidrule(lr){2-5}\cmidrule(lr){6-9}\cmidrule(lr){10-13}
        Method & sparse & medium & dense & global & sparse & medium & dense & global & sparse & medium & dense & global \\
        \midrule
        $N$ detections & $211$ & $4\,541$ & $337\,338$ & $342\,090$ & $2\,508\,395$ & $23\,079\,563$ & $304\,305\,539$ & $329\,893\,497$ & $325$ & $3\,794$ & $40\,438\,584$ & $40\,442\,703$ \\
        \addlinespace[2pt]
        Reference & $-8.128$ & $-4.340$ & $-2.931$ & $-2.953$ & $-1.573$ & $-1.722$ & $-1.550$ & $-1.562$ & $-6.801$ & $-5.036$ & $-2.053$ & $-2.054$ \\
        Sharing & $\mathbf{-7.367}$ & $\mathbf{-4.017}$ & $\mathbf{-2.823}$ & $\mathbf{-2.840}$ & $\mathbf{-1.572}$ & $-1.722$ & $-1.550$ & $\mathbf{-1.561}$ & $\mathbf{-6.274}$ & $\mathbf{-4.787}$ & $-2.053$ & $\mathbf{-2.051}$ \\
        \bottomrule
    \end{tabular*}
\end{table*}

\subsection{Evidence Sharing}
\label{sec:evidence_sharing_ablation}
The sharing of \Cref{sec:evidence_sharing} gives a voxel's neighborhood estimate the weight of $\nu$ crossings, so the neighborhood holds a share $\nu/(C+\nu)$ of the weights it reports and that share falls as its own crossings $C$ accumulate. Because the shrinkage is applied at read time, both settings score the same trained state. We compare $\nu = 3$ (Sharing) against $\nu = 0$ (Reference, each voxel reporting its own mean alone) and report MLPD stratified by per-voxel coverage (\Cref{tab:evidence_sharing}). TBD supplies the sparse embodied regime the mechanism targets; ATC and HB the dense regime in which the shrinkage should be negligible.
Strata are fixed per-voxel detection counts set per dataset: sparse, medium and dense hold below $10$, $10$ to $50$, and above $50$ detections on TBD and HB, and below $5\,000$, $5\,000$ to $20\,000$, and above $20\,000$ on ATC.

The improvement is ordered by coverage on every dataset, as the share $\nu/(C+\nu)$ predicts: $0.761$, $0.323$ and $0.108$ from the sparse to the dense stratum on TBD, and $0.527$ and $0.249$ in HB's two sparse strata against no change in its dense one. 
On ATC the least-covered stratum already contains $2.5$ million detections, resulting in a $0.001$ gain.
The global MLPD understates all of this: HB improves by $0.003$ globally because its two sparse strata hold four thousand of $40.4$ million detections, so a whole-dataset average reports the dense stratum and conceals the $0.527$ gain. Across the three datasets no stratum loses more than $0.0001$, so sharing is enabled by default: it supplies a prior where a voxel's own evidence is thin and withdraws where it is not.

\begin{table*}[t]
    \centering
    \caption{\textbf{Flow dependence ablation} on TBD~\cite{wang2024tbd},
    ATC~\cite{brvsvcic2013person} and HB~\cite{apeltauer2024spatiotemporal}, over
    adjacent-voxel pairs. Results report gains over the uncoupled model
    ($\phi \equiv 0$): $\Delta$J in joint MLPD, $\Delta$MAE in conditional
    heading error, so higher is better for both. The lower block stratifies the
    learned coupling by $D_{\mathrm{KL}}(\boldsymbol{\pi}_{\mathbf{i}} \Vert \boldsymbol{\pi}_{\mathbf{j}})$.
    }
    \label{tab:coupling_ablation}
    \scriptsize
    \setlength{\tabcolsep}{3pt}
    \begin{tabular*}{\textwidth}{@{\extracolsep{\fill}}ll cc cc cc cc cc@{}}
    \toprule
     & & \multicolumn{2}{c}{TBD} & \multicolumn{2}{c}{ATC (emb.)} & \multicolumn{2}{c}{ATC (full)} & \multicolumn{2}{c}{HB (emb.)} & \multicolumn{2}{c}{HB (full)} \\
    \cmidrule(lr){3-4}\cmidrule(lr){5-6}\cmidrule(lr){7-8}\cmidrule(lr){9-10}\cmidrule(lr){11-12}
    Pairs & Coupling & $\Delta$J & $\Delta$MAE & $\Delta$J & $\Delta$MAE & $\Delta$J & $\Delta$MAE & $\Delta$J & $\Delta$MAE & $\Delta$J & $\Delta$MAE \\
    \midrule
    \multirow{2}{*}{\makecell[c]{\emph{All}}}
    & Fixed alignment prior & $\mathbf{\phantom{-}0.25}$ & $\phantom{-}26.2^\circ$ & $\mathbf{\phantom{-}0.29}$ & $\mathbf{\phantom{-}33.5^\circ}$ & $\phantom{-}0.28$ & $\phantom{-}32.9^\circ$ & $\phantom{-}0.42$ & $\phantom{-}51.0^\circ$ & $\phantom{-}0.43$ & $\phantom{-}52.1^\circ$ \\
    & Learned coupling & $\phantom{-}0.02$ & $\mathbf{\phantom{-}30.2^\circ}$ & $\phantom{-}0.27$ & $\phantom{-}31.0^\circ$ & $\mathbf{\phantom{-}0.38}$ & $\mathbf{\phantom{-}35.9^\circ}$ & $\mathbf{\phantom{-}0.63}$ & $\mathbf{\phantom{-}56.5^\circ}$ & $\mathbf{\phantom{-}0.62}$ & $\mathbf{\phantom{-}57.1^\circ}$ \\
    \cmidrule(l{0pt}r{0pt}){1-12}
    \multirow{3}{*}{\makecell[c]{\emph{Learned,}\\\emph{by} $D_{\mathrm{KL}}$}}
    & $D_{\mathrm{KL}} < 0.2$ & $\phantom{-}0.03$ & $\phantom{-}32.4^\circ$ & $\phantom{-}0.28$ & $\phantom{-}32.6^\circ$ & $\phantom{-}0.39$ & $\phantom{-}38.0^\circ$ & $\phantom{-}0.63$ & $\phantom{-}56.6^\circ$ & $\phantom{-}0.62$ & $\phantom{-}57.0^\circ$ \\
    & $0.2 \le D_{\mathrm{KL}} < 1.0$ & $-0.11$ & $\phantom{-}14.0^\circ$ & $\phantom{-}0.05$ & $\phantom{-}8.2^\circ$ & $\phantom{-}0.13$ & $\phantom{-}8.2^\circ$ & $\phantom{-}0.33$ & $\phantom{-}23.5^\circ$ & $\phantom{-}0.20$ & $\phantom{-}22.9^\circ$ \\
    & $D_{\mathrm{KL}} \ge 1.0$ & $-0.11$ & $\phantom{-}3.4^\circ$ & $-0.29$ & $\phantom{-}0.4^\circ$ & $-0.11$ & $-5.4^\circ$ & $-0.32$ & $\phantom{-}12.3^\circ$ & $-0.62$ & $\phantom{-}3.9^\circ$ \\
    \bottomrule
    \end{tabular*}
\end{table*}

\subsection{Flow Dependence}\label{sec:coupling_eval}

The coupling term (\Cref{sec:flow_dependence}) promotes the per-voxel marginals to a joint distribution over adjacent-voxel pairs. Samples are taken at $S = 12$ per cycle of the shortest candidate period, giving a sampling
window $\Delta t^{}_{\phi} = 5$\,s on TBD and $300$\,s on ATC and HB, with $N^{}_{\min} = 2$ pairs per window.

The ablation evaluates the coupling term at three settings: removed ($\phi_{\mathbf{i}\mathbf{j}}^{(k_{\mathbf{i}}, k_{\mathbf{j}})} \equiv 0$, Kairos without coupling), fixed (a prescribed same-slot alignment with no learned parameters), and learned. 
The primary metric is the joint-MLPD gain $\Delta$J over the removed setting: the log-density of the two simultaneous observations under the pair model, against independent-marginal scoring. The secondary metric is the \emph{conditional heading MAE}: the error at voxel $\mathbf{j}$ when its heading is predicted from the simultaneous observation at adjacent voxel $\mathbf{i}$.
Pairs are further stratified by $D^{}_{\mathrm{KL}}(\boldsymbol{\pi}^{}_{\mathbf{i}} \Vert \boldsymbol{\pi}^{}_{\mathbf{j}})$, the Kullback--Leibler divergence between the two voxels' training-time mixing weights, which measures how differently the two cells flow overall: small values mark pairs that share a flow profile.

Coupling in either form improves both metrics on ATC and HB in both regimes, and marginally on TBD, and the two metrics separate what it contributes (\Cref{tab:coupling_ablation}). The conditional heading MAE carries the large effect: predicting a voxel's heading from a simultaneous observation at its neighbor reduces the error by $31.0^\circ$ on ATC and $56.5^\circ$ on HB. The joint gain ranges from $+0.02$ on TBD to $+0.63$ on HB. TBD dissociates the two metrics: its conditional error falls by $30.2^\circ$ while the pair density remains within $0.02$ of the uncoupled reference. Coupling contributes primarily by conditioning one voxel's prediction on its neighbor, and secondarily by the pairwise structure the joint density measures.

Whether to learn the coupling is decided by per-edge evidence: the learned arm beats the prior only where that evidence is dense (as in HB, and ATC at full visibility).
Within ATC the learned increment over the fixed prior grows with observability, from $-0.02$ embodied to $+0.10$ at full visibility. On TBD the learned joint gain is $+0.02$ against the prior's $+0.25$: per-edge estimation requires pair observations that the shortest deployments do not provide, and the parameter-free prior is preferable there. Where evidence is present learning the coupling performs better on both metrics, by $+0.21$ in joint gain and $5.5^\circ$ in conditional MAE on embodied HB.

The $D_{\mathrm{KL}}$ stratification locates the gain, and both metrics concentrate on the same strata: on pairs whose marginals are similar the joint gain is largest ($+0.28$ on ATC embodied at $D_{\mathrm{KL}} < 0.2$, $+0.39$ at full visibility) and so is the error reduction ($32.6^\circ$ and $38.0^\circ$). 
On divergent pairs the joint gain is negative ($-0.29$ and $-0.11$ at $D^{}_{\mathrm{KL}} \ge 1.0$) and the error reduction shrinks ($0.4^\circ$ and $-5.4^\circ$).
HB shows the same ordering ($+0.63$ on similar pairs, $-0.32$ on divergent ones, embodied). 
Coupling is therefore informative where the neighbor's flow resembles the target cell's, and restricting $\phi$ to low-divergence pairs is a direct refinement.

\subsection{Calibration of Predictive Uncertainty}
\label{sec:unc_eval}

\subsubsection{Speed}

The speed channel emits a per-detection interval $\hat{\sigma}^{}_\rho$. In the evaluated configuration the per-slot speed estimate is a running mean, so this interval is the aleatoric variance \eqref{eq:aleatoric} alone: each cell's measured speed scatter, regularized by the prior where evidence is sparse.
We report empirical coverage at the $68\%$ and $95\%$ levels, the Expected Calibration Error (ECE), and sharpness (the mean interval width $2\hat{\sigma}^{}_\rho$).

Coverage is $0.76$ and $0.95$ on ATC and $0.66$ and $0.95$ on HB against nominal $0.68$ and $0.95$, at ECE $0.06$ and $0.01$ (\Cref{tab:unc_ablation}).
The interval widths correspond to a per-detection standard deviation of $0.27$\,m/s on ATC and $0.40$\,m/s on HB, a difference no per-site parameter produces: \eqref{eq:aleatoric} blends each cell's measured scatter with a prior common to both datasets. Coverage at the $95\%$ level is exact at both sites; at $68\%$ the intervals are conservative, the safe direction for a planner reading them. The interval thus follows each environment's scatter without recalibration.

\subsubsection{Weights}

The weight channel emits, per slot, the mixing weight together with its epistemic variance \eqref{eq:epistemic}, propagated through the simplex normalization.
A mixing weight is the fraction of a voxel's crossings belonging to a slot. That fraction is not directly observable, so we compare the predicted weight against the mean soft responsibility \eqref{eq:responsibility} over the crossings falling in a non-overlapping time window at one voxel, rebuilt with the trainer's own crossing definition. A window holds finitely many crossings, so its mean responsibility departs from the voxel's mixing weight by an amount governed by that count. That departure is the channel's aleatoric term, formed per window from that count, and the reported credible interval combines it with the epistemic variance.

\begin{table}[t]
    \centering
    \caption{\textbf{Uncertainty calibration} on ATC~\cite{brvsvcic2013person} 
    and HB~\cite{apeltauer2024spatiotemporal}: coverage, expected calibration error and sharpness. 
    }
    \label{tab:unc_ablation}
    \scriptsize
    \setlength{\tabcolsep}{2pt}
    \begin{tabular*}{\columnwidth}{@{\extracolsep{\fill}}lcccccccc@{}}
        \toprule
         & \multicolumn{4}{c}{ATC} & \multicolumn{4}{c}{HB} \\
        \cmidrule(lr){2-5}\cmidrule(lr){6-9}
        & Cov.\ 68 & Cov.\ 95 & ECE $\downarrow$ & Sharp.\ $\downarrow$ & Cov.\ 68 & Cov.\ 95 & ECE $\downarrow$ & Sharp.\ $\downarrow$ \\
        \midrule
        Speed & $0.76$ & $0.95$ & $0.06$ & $0.54$ & $0.66$ & $0.95$ & $0.01$ & $0.80$ \\
        Weights & $0.71$ & $0.93$ & $0.02$ & $0.24$ & $0.69$ & $0.96$ & $0.01$ & $0.37$ \\
        \bottomrule
    \end{tabular*}
\end{table}

The weight channel is well calibrated on both datasets. Coverage is $0.707$ and $0.935$ on ATC and $0.699$ and $0.964$ on HB against the nominal $0.68$ and $0.95$, at ECE $0.022$ and $0.011$. A cluster bootstrap over $36{,}168$ and $2{,}783$ voxel-day units (the resampling unit since windows sharing pedestrians are not independent samples) places the ECE within $[0.022, 0.023]$ and $[0.010, 0.011]$, against the $0.0002$ and $0.0004$ of a perfect model.
Attributing each crossing to its nearest component rather than splitting it across components, the same intervals cover $0.891$ and $0.922$ at the $95\%$ level.

We decompose the predictive variance by the training evidence a voxel carries, on ATC since HB supports no such stratification. The epistemic term supplies $0.7\%$ of the predictive variance across the $8.6$ million windows above $200$ training crossings, $19.0\%$ across the $12{,}384$ below $200$, and $52.9\%$ across the $2{,}440$ of those falling below $50$, with no per-cell parameter setting the transition: the credible interval reduces to its aleatoric term where the mixing weights are determined by the data and is governed by the epistemic term where they are not. Coverage falls across the same strata, from $0.936$ to $0.866$ and $0.832$ against nominal $0.95$, so the epistemic term widens the interval in the right direction without fully accounting for the miscoverage on thin evidence.
\section{Conclusion}\label{sec:conclusion}

We presented Kairos, a predictive directional-flow memory that extends a hierarchical 3D scene graph to a 4D scene graph. At any traversable location and any future query time, Kairos predicts whether people are likely to be present and how they are likely to be moving, as a full directional distribution anchored to the reconstructed geometry. We evaluated it on three real pedestrian environments, from minutes of robot operation to a station concourse recorded continuously for eleven months, against dedicated occupancy and flow baselines. Kairos attains the best scores in every channel on the robot-collected dataset and is the only compared method that predicts the speed, joint, and presence densities together.
Its presence forecasts are competitive with a dedicated occupancy model trained on the full detection stream, while observing a small fraction of it. Its credible intervals are calibrated, and its learned state remains consistent under loop-closure corrections. On a downstream person-finding task, planning over Kairos's forecasts succeeds more often and encounters more people than planning over any time-invariant map.

\textbf{Limitations.} Three limitations bear on the estimates the system reports. First, the mixing-weight and presence predictors accumulate cumulative means, so every observation carries equal weight: a place whose flow changes permanently is corrected at a rate of one over its accumulated sample count, and no test separates such a change from sampling noise.
Second, the learned coupling requires per-edge observations that short deployments do not supply so the parameter-free prior is preferable there.
Third, the weight credible interval under-covers where training evidence is thin since a Gaussian interval propagated through the simplex normalization does not respect the bounded support of a mixing weight.

\textbf{Future Work.} Each limitation admits a direct refinement. 
Recency weighting, available on the speed channel, extends to the weight and presence predictors, paired with a change test that resets a cell rather than diluting it.
Restricting $\phi$ to low-divergence pairs confines the coupling to where it is informative, and lifting it from adjacent voxels to the place topology would represent a corridor as a single field. A Dirichlet posterior over the slots, formed from the crossing counts the state already carries, replaces the normalized Gaussian interval with one that respects the simplex and widens as those counts fall. 
Finally, motion archetypes are derived from the flow state but not consumed by it; using a place's archetype as the cold-start prior for a new voxel would extend evidence sharing from spatial neighbors to places that behave alike.

\bibliographystyle{IEEEtran}
\bibliography{bibliography}

\clearpage
\appendices

\section{Temporal Structure of Pedestrian Flow}
\label{sec:supplementary}

We report a model-free measurement of how pedestrian activity, heading, and speed vary across six real environments over selected temporal cycles. Its purpose is to establish which of the two quantities Kairos factorizes (presence and the direction mix) varies with time, at which periods, and how strongly.

\subsection{Datasets and Candidate Periods}
\label{sec:supp_datasets}


\begin{table}[t]
  \centering
  \caption{\textbf{Datasets} analyzed in the temporal structure of pedestrian flow study.}
  \label{tab:supp_datasets}
  \footnotesize
  \setlength{\tabcolsep}{3pt}
  \resizebox{\columnwidth}{!}{
  \begin{tabular}{llll}
  \toprule
  Dataset & Environment & Time Span & Rate \\
  \midrule
  ATC~\cite{brvsvcic2013person} & shopping mall & 84 days / $\sim$1 yr & 10 Hz \\
  Eindhoven~\cite{pouw2024data} & train platform & 60 days & 10 Hz \\
  Havl\'i\v{c}k\r{u}v Brod~\cite{apeltauer2024spatiotemporal} & station concourse & 11 months, 24/7 & \multirow{2}{*}{$\sim$1 Hz} \\
  Pardubice~\cite{apeltauer2024spatiotemporal} & station concourse & 11 months, 24/7 & \\
  TBD~\cite{wang2024tbd} & university hall & ${\sim}$213\,min/16 scenes & 10 Hz \\
  Edinburgh~\cite{majecka2009statistical} & university hall & $\sim$7 months & $\sim$9 Hz \\
  \bottomrule
  \end{tabular}
  }
\end{table}

We select candidate periods from the type of environment (\Cref{tab:supp_datasets}) and the temporal coverage of its recordings, following the approach used for the predictors in~\Cref{sec:setup}. ATC~\cite{brvsvcic2013person} and the station concourses  Havl\'i\v{c}k\r{u}v Brod (HB) and Pardubice (PB)~\cite{apeltauer2024spatiotemporal} datasets provide calendar timestamps, allowing daily and weekly folds and separate weekday and weekend analyses. Eindhoven~\cite{pouw2024data} provides a continuous sequence without date labels, allowing repeated short-cycle measurements but limiting calendar-based interpretation. Edinburgh's~\cite{majecka2009statistical} approximate timestamps support analysis at resolutions of five minutes or longer. TBD~\cite{wang2024tbd} contains short sessions, totaling approximately $213$ minutes, and is evaluated at periods of $1$, $5$, and $10$ minutes.

We consider daily and weekly candidate periods at ATC; daily, weekly, morning/afternoon, and $30$- and $60$-minute cycles at HB and PB; a daily and $30$- and $15$-minute cycles at Eindhoven; and a daily period at Edinburgh. At Eindhoven, the $30$-minute period represents the expected timetable cadence and the $15$-minute provides a comparison.

\subsection{Evaluation Protocol}
\label{sec:supp_measurement}

\paragraph{Spatial cells and temporal phases}
We partition each environment into $0.4$\,m cells (Kairos' flow voxel resolution). For a candidate period $T$ we define a \emph{bin resolution} that partitions the phase into \emph{phase bins} of equal width, and each observation is assigned accordingly (\ie{} at hourly resolution of a daily period, each bin contains the observations recorded at the same hour across days). We bin under the assumption that the distributions below are roughly constant across a bin width.

\paragraph{Observation units and measurements}
We select moving detections with speed above $0.5$\,m/s. The observation unit is the \emph{passage of a cell}. At $10$\,fps a single crossing yields tens of near-identical detections and a lingering agent thousands, so counting detections would conflate time spent in a cell with the number of agents crossing it. We therefore retain one observation per track, cell, and phase. We define a cell's \emph{activity} in a phase bin as the number of observations lying in it. We assign each observation's heading to one of $K$ \emph{heading bins} (eight, one per cardinal direction). For each cell and each phase bin we then form the normalized histogram over those bins and over binned speed. We also measure the fraction of activity within $\pm45^\circ$ of the direction opposite the cell's dominant heading, which distinguishes predominantly one-way from bidirectional locations.

For heading and speed we retain moving detections with speed above $0.5$\,m/s, and their observation unit is the \emph{passage of a cell}. At $10$\,fps a single crossing yields tens of near-identical detections and a lingering agent thousands, so counting detections would conflate time spent in a cell with the number of agents crossing it. We assign each detection's heading to one of $K$ \emph{heading bins} (eight, one per cardinal direction) and its speed to one of six bins of equal width over $[0.5, 3.0]$\,m/s, and retain one observation per track, cell, phase bin, and value bin, so a passage whose detections span two heading bins contributes to both. A cell's \emph{traffic} is the number of observations retained there, counted separately for heading and for speed. For each cell and each phase bin we form the normalized histogram over the heading and speed bins.
We define \emph{activity} the count every detection at the source frame rate, stationary people included.
We also measure the fraction of a cell's traffic assigned to the heading bin opposite its dominant one and the two bins adjacent to it, pooled over all phase bins, which distinguishes predominantly one-way from bidirectional locations.

\paragraph{Signal and reference divergence}
For heading and speed, we compare the cell's distribution in each phase bin with its distribution pooled across phases, using Jensen--Shannon (JS) divergence with base-two logarithms, and average the divergences over phase bins, weighting each by its share of the cell's traffic. The resulting \textbf{signal} measures how much that distribution changes with phase. For activity, the signal is instead the divergence between the cell's normalized activity profile and a uniform profile.
We divide the recording into alternating days, weeks, or cycles, as appropriate for the candidate period. For heading and speed, we take the JS divergence between the two halves' distributions at each phase bin and average it as above; for activity, we take the divergence between the two halves' phase profiles, each normalized to sum to one. The resulting \textbf{reference divergence} measures disagreement between repeated observations of the same phase, including finite-sample variation, differences between days or cycles, and any dependence remaining within the observations.
We report the signal divided by the reference divergence as S/R. Since both are computed from finite counts, neither reaches zero even when the underlying distributions agree.
The signal compares a phase bin against a marginal pooled over every bin, whereas the reference compares two halves of that one bin, each holding half its observations. The reference therefore has up to four times the sampling variance of the signal, and a cell with no phase-dependent variation yields between $0.25$ and $0.5$ rather than $1$, approaching the lower end as the number of phase bins grows.
We take $0.3$ as that null, and read a ratio above $3$ as variation beyond what repeated observation produces. A coarser division of the period pools more observations into each bin and lowers the reference divergence, so it detects more at the cost of any variation that changes within a bin.

We also report a site-level ratio: the mean signal over the $200$ cells with the most observations divided by their mean reference divergence, weighted by those same counts.

\subsection{Variation in Activity}
\label{sec:supp_presence}

Activity varies most with hour of day at HB and PB, where all $200$ analyzed cells at each station exceed S/R $3$, with a median per-cell S/R near $150$. ATC, Eindhoven, and Edinburgh also vary with the hour, at $22$, $6.0$, and $14.2$. Activity also varies with day of week at ATC ($24$) and weakly at HB ($4.4$), while PB stays below the threshold ($1.1$). At Eindhoven it varies within the $30$-minute period as well, at S/R $5.0$, changing by about a factor of $4.8$ between the quietest and busiest phases. The hour-of-day results use weekdays only at HB, PB, and Edinburgh, and all recorded days at ATC.

The activity divergence is $0.34$ bits at ATC and about $0.13$ bits at HB and PB. ATC's is the larger despite its much lower S/R, because ATC is recorded only during opening hours and its daily profile therefore includes hours of zero, while HB and PB are recorded continuously.

\subsection{Variation in Heading}
\label{sec:supp_direction}

\begin{figure}[t]
  \centering
  \includegraphics[width=\columnwidth]{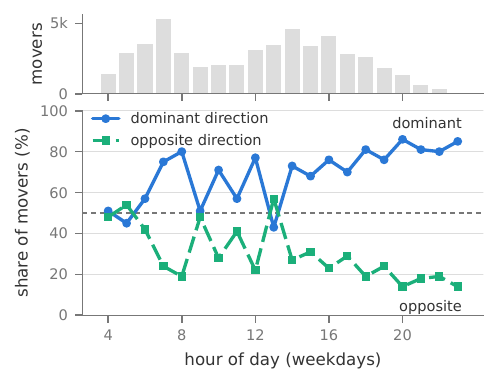}
  \caption{\textbf{Hourly heading profile} at the HB cell selected for its strong variation, restricted to weekdays and hours with at least $20$ movers. \textbf{Top}: movers per hour. \textbf{Bottom}: shares within $\pm45^\circ$ of the cell's dominant and opposite headings.}
  \label{fig:supp_reversal}
\end{figure}

\textbf{HB.} At HB on weekdays the site-level S/R is $2.5$, below the threshold, on a signal of $0.011$ bits. Per cell, the median S/R is $2.1$, while $49$ of the $200$ cells exceed the threshold and hold $34\%$ of the traffic. These cells have an S/R of $4$--$9$ and a heading signal of $0.007$--$0.018$ bits: detectable, but far below the $0.13$ bits of activity variation at the same station. Heading therefore varies with hour of day in a quarter of the cells and not in the site-wide measurement. Dividing the day into morning and afternoon instead puts $60$ cells above the threshold, while at periods of $30$ and $60$ minutes the site-level S/R is $0.3$ and $1.1$.

A cell's S/R averages over the day, so individual hours can depart from the cell's overall distribution by more than it indicates. \Cref{fig:supp_reversal} shows the hourly profile of a cell selected for its strong variation. Its dominant heading, fixed over the whole recording, holds $86\%$ of the traffic at some hours and $45\%$ at others: nearly one-way at one time of day, close to balanced at another, with the reverse flow strongest in the early-commuter and midday hours.

\textbf{PB.} In contrast, PB shows no detectable variation by hour of day and clear variation between morning and afternoon. Its hourly S/R is $1.2$, below the threshold at every cell; dividing the day into morning and afternoon instead puts $86$ of its $200$ cells above it. At periods of $30$ and $60$ minutes the S/R is $0.4$ and $0.6$.

\textbf{ATC.} At ATC the site-level S/R is approximately $15$ and $197$ of the $200$ cells exceed the threshold, accounting for $98\%$ of the traffic. The signal is $0.0099$ bits against a reference divergence of approximately $0.0007$ bits. By day of week the signal is lower, at $0.0025$ bits and S/R $6.4$. In several busy corridor cells the share of traffic running opposite the dominant heading is $47$--$56\%$ in the first recorded hour, against $28$--$35\%$ later in the day, so these cells are close to balanced early and one-way afterwards.

\textbf{Eindhoven.} Eindhoven varies within the $30$-minute period and not by hour of day: S/R $8.5$ against $1.45$. Within that period $141$ of the $200$ cells exceed the threshold and account for $55\%$ of the traffic.
At bidirectional mid-platform cells the along-platform share runs from about $15\%$ to $80\%$, while predominantly one-way corridors vary in activity without a comparable change in heading. At a $15$-minute period the S/R is $0.24$, since folding a $30$-minute rhythm at half its length superimposes the two halves of each cycle. Eindhoven and the concourses are therefore opposite cases: the period that reveals structure at one shows none at the other.

\textbf{Edinburgh and TBD.} Neither presents variation with hour of day with an S/R of $0.6$ and $0.2$--$0.3$, and no cell above the threshold.

\subsection{Variation in Speed}
\label{sec:supp_speed}

Speed varies with hour of day at ATC, where the signal is $0.006$ bits at S/R $15.9$ and $199$ of the $200$ cells exceed the threshold. At HB the site-level S/R is $1.6$, below the threshold, on a signal of $0.004$ bits, with $13$ cells above it. At TBD the S/R is $0.2$, with no cell above the threshold.

The variation is detectable but small. At ATC's cells with the most observations the mean speed changes by $0.02$--$0.06$\,m/s across the day, with roughly stable dispersion, against the $0.3$\,m/s standard deviation of the model's speed components. Speed also has the smallest divergence of the three quantities, below heading's $0.0099$ bits at the same site.

\subsection{Interpretation and Model Design}
\label{sec:supp_why}
\label{sec:supp_consequences}

\begin{figure}[t]
  \centering
  \includegraphics[width=\columnwidth]{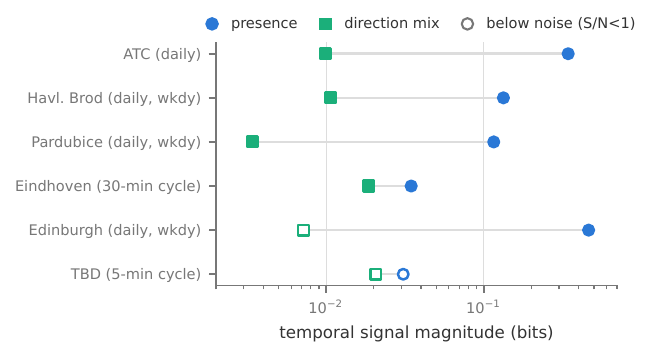}
  \caption{\textbf{Temporal divergences} for activity and heading at each site's primary period. Open markers denote aggregate S/N below $1$.}
  \label{fig:supp_magnitudes}
\end{figure}

Activity and heading describe different aspects of pedestrian motion: activity can increase while directional proportions remain unchanged, and activity divergences exceed heading and speed divergences throughout (\Cref{fig:supp_magnitudes}). The measurements therefore support estimating presence separately from the conditional flow distribution.

Directional changes are concentrated at particular locations and phases rather than spread evenly through the period, which supports selecting the spectral order per cell.

Repeated detections also require appropriate observation units: the heading estimator uses crossings, while the presence estimator accounts separately for visible time and dwell.

The hourly speed changes are small against the width of the model's speed components, which supports holding the per-slot speed means time-invariant.

\section{Heading and Speed Dependence Within a Component}
\label{sec:supp_covariance}


\begin{table}[t]
  \centering
  \caption{\textbf{Heading and speed dependence} within and across slots.}
  \label{tab:supp_covariance}
  \setlength{\tabcolsep}{4pt}
  \begin{tabular*}{\columnwidth}{@{\extracolsep{\fill}}lcccc@{}} 
  \toprule
  & \multicolumn{3}{c}{within slot} & \multicolumn{1}{c}{across slots} \\
  \cmidrule(lr){2-4} \cmidrule(lr){5-5}
  Dataset & med $r$ & $r$ 5th/95th & $r^2$ & $\eta^2$ \\
  \midrule
  TBD & $-0.018$ & $-0.49$ / $0.47$ & $0.058$ & $0.181$ \\
  ATC & $-0.003$ & $-0.25$ / $0.26$ & $0.019$ & $0.070$ \\
  HB  & $\phantom{-}0.009$ & $-0.42$ / $0.43$ & $0.057$ & $0.186$ \\
  \bottomrule
  \end{tabular*}
\end{table}

Kairos uses diagonal component covariances and a separate mean speed for each directional slot. Within one slot, a detection's heading is at most $\pi/8$ from the slot centre; we measure the linear heading--speed dependence within slots with the Pearson correlation $r$ between that offset and the detection's speed. We evaluate whether speed differs between directions with the correlation ratio $\eta^2$. Each detection is assigned to a single nearest slot, where the model spreads it across neighbouring components.

\textbf{Within slots.} The dependence between a detection's offset and its speed has no consistent sign.
The median correlation over all voxels and slots is $-0.018$ on TBD, $-0.003$ on ATC, and $+0.009$ on HB, whereas the middle $90\%$ of correlations spans $-0.49$ to $0.47$ on TBD, $-0.25$ to $0.26$ on ATC, and $-0.42$ to $0.43$ on HB (\Cref{tab:supp_covariance}). The dependence is therefore \emph{local}, taking one sign in some voxels and slots and the opposite in others.
Moreover, its magnitude is small: the offset accounts for $5.8\%$ of within-slot speed variance on TBD, $1.9\%$ on ATC, and $5.7\%$ on HB. The component covariance is a single matrix shared by every voxel (\Cref{sec:flow_model}), so a cross-covariance term would take one value throughout the map. The value minimizing the error across all of them is close to zero, which is what the diagonal covariance already specifies.

\textbf{Across slots.} Speed differs between directions: $\eta^2$ is $0.181$ on TBD, $0.070$ on ATC, and $0.186$ on HB. We represent these differences through per-slot mean speeds.

\section{Temporal Forecasting of Direction Weights}
\label{sec:supp_error}

\subsection{Contribution of Temporal Weights on ATC}

\begin{table*}[t]
    \centering
    \caption{\textbf{Temporal forecasting of the direction mixing weights} on ATC~\cite{brvsvcic2013person}: change in joint MLPD from replacing static weights (order-0) with the gated NUDFT forecast (order-1), by hour of day.
    {Entries are in units of} ${10^{-3}}$ {nats}; positive favors the forecast.
    \emph{All} is the whole test set.}
    \label{tab:temporal_atc}
    \scriptsize
    \setlength{\tabcolsep}{3pt}
    \begin{tabular*}{\textwidth}{@{\extracolsep{\fill}}l*{13}{c}@{}}
        \toprule
        Regime & 9 & 10 & 11 & 12 & 13 & 14 & 15 & 16 & 17 & 18 & 19 & 20 & All \\
        \midrule
        Embodied ($5$\,m FoV) & $18.5$ & $19.5$ & $3.6$ & $0.6$ & $10.4$ & $7.3$ & $4.3$ & $4.0$ & $13.8$ & $5.5$ & $11.7$ & $7.6$ & $\mathbf{7.6}$ \\
        Full visibility & $16.4$ & $22.7$ & $8.9$ & $4.5$ & $14.7$ & $10.2$ & $6.7$ & $7.3$ & $16.7$ & $8.4$ & $20.8$ & $24.0$ & $\mathbf{11.5}$ \\
        \bottomrule
    \end{tabular*}
\end{table*}

We compare Kairos with its static-weight variant on the same held-out observations, keeping the component speed model fixed. This isolates the effect of forecasting the mixing weights on joint MLPD.

\begin{figure}[t]
  \centering
  \includegraphics[width=\columnwidth]{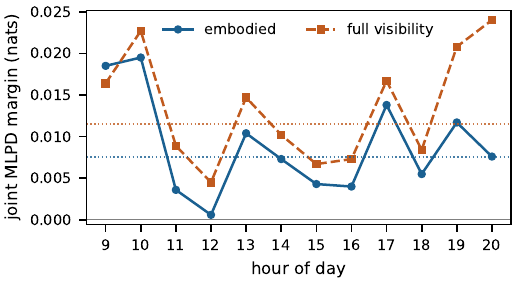}
  \caption{\textbf{Joint-MLPD difference} between temporal and static weights on ATC in both regimes (\emph{embodied} and \emph{full visibility}).}
  \label{fig:hourly_margin}
\end{figure}

On ATC, temporal weights improve joint MLPD by $0.0076$ nats in the embodied regime and $0.0115$ at full visibility, with every reported hour positive (\Cref{tab:temporal_atc}). The gain varies by hour (\Cref{fig:hourly_margin}): largest at 09:00--10:00 ($0.0185$--$0.0195$ nats) and smallest at 12:00 ($0.0006$). At full visibility the largest difference is $0.0240$ nats at 20:00. HB's aggregate gains are $0.0117$ and $0.0139$ nats (\Cref{tab:temporal}).

\subsection{Heading Difference with STeF-Map}

We examine whether STeF-Map's heading advantage comes from Kairos's fixed-width angular kernel, which spreads probability away from the dominant directions (\Cref{sec:flow_results}). For each detection we evaluate the predicted probability in the $45^\circ$ sector containing its observed heading, where an uninformative uniform forecast assigns $0.125$ to every sector.

Both methods place less than $1/8$ on the observed sector for a similar minority of detections: $23.7\%$ for Kairos and $21.7\%$ for STeF-Map on ATC, $17.8\%$ overlapping, and $19.9\%$ against $17.3\%$ on HB, $15.2\%$ overlapping. Where only Kairos places less than $1/8$, its median sector mass is $0.114$ on ATC and $0.115$ on HB, against the uniform $0.125$.

Kairos spreads probability across neighboring directions, both when assigning an observation to a component and when evaluating the component densities: at $\sigma^{}_\theta = 0.4$\,rad about a third of a component's mass is assigned outside the $45^\circ$ sector centered on it. Sectors carrying less than $5\%$ of a cell's traffic receive mean masses of $0.087$ from Kairos and $0.078$ from STeF-Map on ATC, against the $0.028$ they actually contain, and $0.088$ and $0.076$ on HB against $0.031$. On HB, Kairos's per-detection MLPD is $0.14$ nats above STeF-Map's for detections below $0.25$\,m/s and $0.08$--$0.15$ nats below it between $0.5$ and $2$\,m/s. A heading measured on a near-stationary pedestrian is dominated by noise and is uncorrelated with the cell's flow, where spread mass loses less than a sharp peak; above $0.5$\,m/s headings are reliable and fall in the dominant directions.

On ATC the mean paired CRPS difference is below $0.001$\,rad; on HB it is $0.002$\,rad. Integrating Kairos's forecast into the same $45^\circ$ sectors before scoring changes its CRPS by less than $0.002$\,rad on ATC and less than $0.001$\,rad on HB, showing two methods are indistinguishable. STeF-Map's advantage on heading MLPD is therefore specific to the log score, which is evaluated at the observed heading alone, whereas circular CRPS penalizes mass in proportion to its angular distance from the observation. Its histogram concentrates mass on the directions carrying more traffic, while Kairos's fixed-width components retain mass on the remaining directions. The tradeoff is between sharpness on common directions and coverage of uncommon ones, which motivates cell-specific angular widths.

\section{Presence Estimation}

\subsection{Presence Estimation by Hour of Day at HB}
\label{sec:supp_presence_hourly}

\begin{table*}[t]
  \centering
  \caption{\textbf{Presence by hour of day} on HB at the $600$\,s horizon: mean log predictive probability (MLPP) per hour, with each hour's occupancy rate for context. The aggregate table of the main text summarizes these columns; the orderings are the same at the other horizons.}
  \label{tab:supp_presence_hourly}
  \scriptsize
  \setlength{\tabcolsep}{1.5pt}
  \resizebox{\textwidth}{!}{
  \begin{tabular}{lcccccccccccccccccccccccc}
  \toprule
  Hour & 00 & 01 & 02 & 03 & 04 & 05 & 06 & 07 & 08 & 09 & 10 & 11 & 12 & 13 & 14 & 15 & 16 & 17 & 18 & 19 & 20 & 21 & 22 & 23 \\
  \midrule
  \multicolumn{25}{l}{\emph{Embodied ($5$\,m field of view)}} \\
    Constant base rate & $\mathbf{-0.28}$ & $\mathbf{-0.28}$ & $\mathbf{-0.29}$ & $\mathbf{-0.33}$ & $-0.45$ & $-0.63$ & $-0.69$ & $-0.76$ & $-0.66$ & $-0.71$ & $-0.68$ & $-0.72$ & $-0.72$ & $-0.76$ & $-0.76$ & $-0.77$ & $-0.75$ & $-0.71$ & $-0.63$ & $-0.58$ & $-0.48$ & $-0.43$ & $\mathbf{-0.37}$ & $-0.34$ \\
    Kairos (static rate) & $-0.60$ & $-0.60$ & $-0.59$ & $-0.56$ & $-0.50$ & $-0.51$ & $\mathbf{-0.52}$ & $\mathbf{-0.53}$ & $\mathbf{-0.48}$ & $\mathbf{-0.49}$ & $\mathbf{-0.48}$ & $\mathbf{-0.50}$ & $\mathbf{-0.48}$ & $\mathbf{-0.53}$ & $\mathbf{-0.51}$ & $\mathbf{-0.51}$ & $-0.50$ & $\mathbf{-0.48}$ & $\mathbf{-0.46}$ & $-0.45$ & $-0.47$ & $-0.50$ & $-0.53$ & $-0.56$ \\
    Kairos & $-0.32$ & $-0.33$ & $-0.36$ & $-0.39$ & $\mathbf{-0.42}$ & $\mathbf{-0.50}$ & $\mathbf{-0.52}$ & $\mathbf{-0.53}$ & $-0.50$ & $-0.50$ & $-0.50$ & $-0.51$ & $-0.50$ & $\mathbf{-0.53}$ & $\mathbf{-0.51}$ & $\mathbf{-0.51}$ & $\mathbf{-0.49}$ & $\mathbf{-0.48}$ & $\mathbf{-0.46}$ & $\mathbf{-0.44}$ & $\mathbf{-0.42}$ & $\mathbf{-0.40}$ & $\mathbf{-0.37}$ & $\mathbf{-0.33}$ \\
  \midrule
  \multicolumn{25}{l}{\emph{Full visibility}} \\
    Constant base rate & $-0.28$ & $-0.28$ & $-0.29$ & $-0.33$ & $-0.45$ & $-0.62$ & $-0.68$ & $-0.75$ & $-0.65$ & $-0.70$ & $-0.67$ & $-0.71$ & $-0.70$ & $-0.75$ & $-0.75$ & $-0.75$ & $-0.73$ & $-0.70$ & $-0.62$ & $-0.57$ & $-0.48$ & $-0.43$ & $-0.37$ & $-0.34$ \\
    FreMEn occupancy & $\mathbf{-0.04}$ & $\mathbf{-0.02}$ & $\mathbf{-0.08}$ & $\mathbf{-0.17}$ & $\mathbf{-0.32}$ & $\mathbf{-0.48}$ & $-0.51$ & $-0.51$ & $\mathbf{-0.47}$ & $\mathbf{-0.47}$ & $-0.47$ & $\mathbf{-0.47}$ & $\mathbf{-0.46}$ & $\mathbf{-0.48}$ & $\mathbf{-0.47}$ & $\mathbf{-0.47}$ & $\mathbf{-0.47}$ & $-0.47$ & $\mathbf{-0.44}$ & $\mathbf{-0.42}$ & $\mathbf{-0.35}$ & $\mathbf{-0.31}$ & $\mathbf{-0.23}$ & $\mathbf{-0.21}$ \\
    Kairos (static rate) & $-0.59$ & $-0.60$ & $-0.59$ & $-0.55$ & $-0.49$ & $-0.50$ & $-0.51$ & $-0.52$ & $\mathbf{-0.47}$ & $-0.48$ & $\mathbf{-0.46}$ & $-0.48$ & $-0.47$ & $-0.51$ & $-0.50$ & $-0.50$ & $-0.48$ & $\mathbf{-0.46}$ & $\mathbf{-0.44}$ & $-0.44$ & $-0.46$ & $-0.49$ & $-0.53$ & $-0.55$ \\
    Kairos & $-0.25$ & $-0.26$ & $-0.29$ & $-0.34$ & $-0.39$ & $\mathbf{-0.48}$ & $\mathbf{-0.50}$ & $\mathbf{-0.50}$ & $-0.49$ & $-0.48$ & $-0.49$ & $-0.49$ & $-0.48$ & $-0.50$ & $-0.49$ & $-0.48$ & $\mathbf{-0.47}$ & $\mathbf{-0.46}$ & $\mathbf{-0.44}$ & $-0.43$ & $-0.39$ & $-0.36$ & $-0.31$ & $-0.27$ \\
  \bottomrule
  \end{tabular}}
\end{table*}

We examine presence MLPP at the $600$\,s event window by hour of day, using the protocol of \Cref{sec:presence_eval}.

In the embodied regime the temporal model's MLPP is $-320$ at 00:00 against the static rate's $-600$, and $-330$ at 23:00 against $-560$; the occupancy rate in those hours is $0.00$ and $0.05$. The advantage narrows to $10$ at 05:00, and from 08:00 through 12:00 the static rate leads by $10$ to $20$. Over the whole day the values are $-460$ and $-508$ (\Cref{tab:presence}). The gain is produced by the hours in which the cell is almost never occupied, where a rate held constant across the day over-predicts activity; through the busiest hours the two are within $20$ of each other.

\subsection{Presence Estimation on ATC}
\label{sec:supp_presence_atc}

\begin{table*}[t]
    \centering
    \caption{\textbf{Presence estimation on ATC}~\cite{brvsvcic2013person} across event windows $\Delta t$: MLPP (in nats), {reliability} (Rel), and {resolution} (Res); all in units of ${10^{-3}}$.
    Reliability and resolution are shown from $60$\,s onward, the windows in which they separate the estimators.
    Rows and units match \Cref{tab:presence}: \emph{Kairos} is the full gated estimator and \emph{Kairos (static rate)} the variant holding the rate time-invariant.
    }
    \label{tab:supp_presence_atc}
    \scriptsize
    \setlength{\tabcolsep}{4pt}
    \begin{tabular*}{\textwidth}{@{\extracolsep{\fill}}cl ccccccccccc@{}}
        \toprule
        & & $\Delta t = 5$\,s & $\Delta t = 10$\,s & \multicolumn{3}{c}{$\Delta t = 60$\,s} & \multicolumn{3}{c}{$\Delta t = 300$\,s} & \multicolumn{3}{c}{$\Delta t = 600$\,s} \\
        \cmidrule(lr){3-3}\cmidrule(lr){4-4}\cmidrule(lr){5-7}\cmidrule(lr){8-10}\cmidrule(lr){11-13}
        Regime & Model & MLPP$\uparrow$ & MLPP$\uparrow$ & MLPP$\uparrow$ & Rel$\downarrow$ & Res$\uparrow$ & MLPP$\uparrow$ & Rel$\downarrow$ & Res$\uparrow$ & MLPP$\uparrow$ & Rel$\downarrow$ & Res$\uparrow$ \\
        \midrule
        \multirow{5}{*}{\makecell[c]{\emph{Embodied}\\\emph{$5$\,m FoV}}}
        & Constant base rate       & $-207$ & $-299$ & $\mathbf{-579}$ & $3.0$ & $0.0$ & $\mathbf{-698}$ & $3.4$ & $0.0$ & $\mathbf{-693}$ & $2.9$ & $0.0$ \\
        & Kairos (no dwell)        & $-218$ & $-362$ & $-1113$ & $61.7$ & $0.2$ & $-2134$ & $211.7$ & $0.3$ & $-2562$ & $294.7$ & $0.4$ \\
        & Kairos (static rate)     & $\mathbf{-192}$ & $\mathbf{-285}$ & $-729$ & $74.4$ & $50.9$ & $-1464$ & $81.4$ & $87.6$ & $-1719$ & $63.7$ & $98.0$ \\
        & Kairos                   & $-202$ & $-303$ & $-806$ & $63.1$ & $51.4$ & $-1597$ & $65.6$ & $83.3$ & $-1887$ & $54.5$ & $89.6$ \\
        \midrule
        \multirow{5}{*}{\makecell[c]{\emph{Full}\\\emph{visibility}}}
        & Constant base rate       & $-195$ & $-283$ & $-557$ & $3.3$ & $0.0$ & $-693$ & $4.4$ & $0.0$ & $-701$ & $4.0$ & $0.0$ \\
        & FreMEn~\cite{krajnik2017fremen} occupancy$\dagger$ & $-197$ & $-492$ & $-1478$ & $57.5$ & $76.0$ & $-1948$ & $45.6$ & $134.1$ & $-1745$ & $35.8$ & $149.7$ \\
        & Kairos (no dwell)        & $-302$ & $-508$ & $-1562$ & $56.3$ & $0.0$ & $-2948$ & $186.7$ & $0.0$ & $-3516$ & $258.2$ & $0.0$ \\
        & Kairos (static rate)     & $\mathbf{-191}$ & $\mathbf{-273}$ & $\mathbf{-482}$ & $31.1$ & $51.7$ & $\mathbf{-489}$ & $23.4$ & $104.6$ & $\mathbf{-459}$ & $12.4$ & $122.0$ \\
        & Kairos                   & $-195$ & $-283$ & $-570$ & $29.3$ & $53.7$ & $-866$ & $30.4$ & $97.3$ & $-1042$ & $27.4$ & $105.2$ \\
        \bottomrule
    \end{tabular*}
    \par\vspace{2pt}%
    \noindent\parbox{\textwidth}{\scriptsize\textit{Notes}:  $\dagger$ FreMEn divides each cell's occupied time by total elapsed time, so time in which the cell was not observed is indistinguishable from time in which it was empty. It is therefore fit only under \emph{full visibility}.
    }
\end{table*}

ATC is recorded during opening hours only, so its record omits the closed part of the daily cycle (\Cref{sec:presence_results}). The baselines follow \Cref{sec:presence_baselines}. FreMEn uses a full-visibility fit, without per-cell order selection.

In the embodied regime MLPP is $-192$ for the static-rate variant and $-202$ for temporal Kairos at $5$\,s, and $-1719$ and $-1887$ at $600$\,s; at full visibility the $600$\,s pair is $-459$ and $-1042$. The static variant is ahead at every window in both regimes, and the deficit grows from $10$ to $168$ embodied and to $583$ at full visibility. When no event occurs in the window, the score for that query is $-\hat{\lambda}\Delta t$ nats \eqref{eq:voxel-presence}. The cost of a given rate error therefore grows with the window duration.

The constant base rate has the best embodied MLPP from $60$\,s onward, $-693$ at $600$\,s against $-1719$ for the best cell-level variant, while at $5$ and $10$\,s the cell-level estimates are better in both regimes. At full visibility the ordering reverses and the static-rate variant exceeds the base rate by $75$ at $60$\,s and $242$ at $600$\,s. The constant base rate estimates one rate from the whole record, so its error does not depend on how often an individual cell was observed. The per-cell rate supports the forecast at short windows under either coverage, and at long windows only under full visibility.

At $600$\,s FreMEn has the highest resolution in the table, $138.3$ embodied and $149.7$ at full visibility, with MLPP of $-1896$ and $-1745$. Removing the dwell correction gives resolution of at most $0.4$ and reliability of $294.7$. 

On HB the gated temporal model has the best embodied MLPP at every window, $-460$ against $-508$ for its static variant at $600$\,s (\Cref{tab:presence}); on ATC that ordering reverses at every window. HB is recorded continuously over eleven months, ATC over $84$ days spread across a year (\Cref{tab:supp_datasets}). What limits the presence forecast on ATC is the rate estimate rather than the temporal model. A per-cell window limit, or a fallback to the global rate in cells with few observations, would address that limit directly.

\section{Predictive Uncertainty}
\label{sec:supp_uncertainty}

\begin{table}[t]
    \centering
    \caption{\textbf{Weight interval by training evidence.} Windows are stratified by the number of crossings the voxel accumulated during training.
    }
    \label{tab:supp_unc_evidence}
    \footnotesize
    \setlength{\tabcolsep}{4pt}
    \begin{tabular*}{\columnwidth}{@{\extracolsep{\fill}}l cccc@{}}
        \toprule
        Dataset & Crossings & Windows & Epi.\ share & Cov.\ 95 \\
        \midrule
        ATC~\cite{brvsvcic2013person} & $\ge 200$ & $8{,}605{,}880$ & $0.007$ & $0.936$ \\
                                      & $< 200$   & $12{,}384$      & $0.190$ & $0.866$ \\
                                      & $< 50$    & $2{,}440$       & $0.529$ & $0.832$ \\
        \midrule
        HB~\cite{apeltauer2024spatiotemporal} & $\ge 200$ & $2{,}963{,}424$ & $0.002$ & $0.964$ \\
                                              & $< 200$   & $704$           & $0.048$ & $0.939$ \\
                                              & $< 50$    & $64$            & $0.145$ & $0.969$ \\
        \bottomrule
    \end{tabular*}
\end{table}

\begin{table}[t]
    \centering
    \caption{\textbf{Speed uncertainty: adapted variance against a constant prior} on ATC~\cite{brvsvcic2013person} and HB~\cite{apeltauer2024spatiotemporal}. The constant row fixes $\hat{\sigma}^{}_\rho$ to the aleatoric prior $\sigma^{}_\rho = 0.3$\,m/s for every detection. Bold marks the better value per column.}
    \label{tab:supp_unc_speed}
    \footnotesize
    \setlength{\tabcolsep}{2pt}
    \resizebox{\columnwidth}{!}{
    \begin{tabular}{l cccc cccc}
        \toprule
         & \multicolumn{4}{c}{ATC~\cite{brvsvcic2013person}} & \multicolumn{4}{c}{HB~\cite{apeltauer2024spatiotemporal}} \\
        \cmidrule(lr){2-5}\cmidrule(lr){6-9}
        & Cov.\ 68 & Cov.\ 95 & ECE $\downarrow$ & Sharp.\ $\downarrow$ & Cov.\ 68 & Cov.\ 95 & ECE $\downarrow$ & Sharp.\ $\downarrow$ \\
        \midrule
        Constant aleatoric prior & $0.79$ & $0.96$ & $0.09$ & $\mathbf{0.60}$ & $0.55$ & $0.84$ & $0.08$ & $\mathbf{0.60}$ \\
        Adapted aleatoric variance & $\mathbf{0.76}$ & $\mathbf{0.95}$ & $\mathbf{0.06}$ & $0.54$ & $\mathbf{0.66}$ & $\mathbf{0.95}$ & $\mathbf{0.01}$ & $0.80$ \\
        \bottomrule
    \end{tabular}
    }
\end{table}

\textbf{Weight Intervals.} The weight forecast is evaluated against the mean soft responsibility of crossings observed within a voxel-time window. Its uncertainty has two sources: uncertainty in the predicted mean weight, and variation in the finite set of crossings used to evaluate it.

Let $p$ be the predicted weight and $v$ its propagated epistemic variance. For $0<p<1$ and $0<v<p(1-p)$, a moment-matched Beta distribution has parameters
\begin{equation}
  a=p\kappa,\qquad b=(1-p)\kappa,
  \qquad \kappa=\frac{p(1-p)}{v}-1.
\end{equation}
The variance is capped below the Bernoulli bound before matching. Conditional on a latent share $P\sim\mathrm{Beta}(a,b)$, an integer count of slot assignments among $n$ independent crossings satisfies $X\mid P\sim\mathrm{Binomial}(n,P)$. Marginalizing $P$ gives the beta-binomial predictive distribution, whose share variance is
\begin{equation}
  \operatorname{Var}(X/n)
  =\frac{p(1-p)}{n}+\left(1-\frac{1}{n}\right)v.
\end{equation}
The first term is the finite-window variation and the second the uncertainty in the mean.

For an integer observation $x$, the randomized probability integral transform is $F(x^-)+U\Pr(X=x)$, where $U$ is uniform on $[0,1]$ and seeded per window and slot. Under a correctly specified count distribution the transformed values are uniform. The evaluated target is a sum of fractional responsibilities, so the beta-binomial is a count-based approximation and the uniformity holds only approximately. The propagation omits cross-slot covariance, although the responsibilities sum to one, and crossings within a voxel-day can be correlated; the reported intervals resample voxel-day units.

Below $200$ training crossings HB has $704$ evaluation windows, and below $50$ it has $64$, against ATC's $12{,}384$ and $2{,}440$ (\Cref{tab:supp_unc_evidence}). Its epistemic shares in those strata are $0.048$ and $0.145$, and its coverage is $0.939$ and $0.969$ against the nominal $0.95$. These strata are too small to support the comparison reported on ATC in \Cref{sec:unc_eval}.

\textbf{Speed Intervals.} Sharpness is the mean interval width $2\hat\sigma^{}_\rho$. With the constant $0.3$\,m/s prior, coverage is $0.79$ and $0.96$ on ATC and $0.55$ and $0.84$ on HB, against the nominal $0.68$ and $0.95$ (\Cref{tab:supp_unc_speed}). With the adapted variance it is $0.76$ and $0.95$ on ATC and $0.66$ and $0.95$ on HB. The mean width decreases from $0.60$ to $0.54$\,m/s on ATC and increases from $0.60$ to $0.80$\,m/s on HB. The measured scatter is $0.27$\,m/s on ATC and $0.40$\,m/s on HB, on either side of the $0.3$\,m/s prior. The estimator therefore corrects the interval in opposite directions at the two sites from the same prior value. A fixed width fits one site or the other, not both.

\end{document}